\pdfoutput=1 

\documentclass[10pt,twocolumn,letterpaper]{article}

\usepackage[pagenumbers]{wacv} 

\newcommand{\cmark}{\textcolor{green!70!black}{\checkmark}}
\newcommand{\xmark}{\textcolor{red}{$\boldsymbol{\times}$}}

\usepackage{graphicx}
\usepackage{tabularx}
\graphicspath{{figs/}{./}}
\usepackage{booktabs}
\usepackage{amsmath}
\usepackage{amssymb}
\usepackage{xspace}
\usepackage[table]{xcolor}
\usepackage{siunitx}
\usepackage{multirow}
\usepackage[normalem]{ulem}   
\usepackage{float}
\usepackage{flafter}

\usepackage{tikz}
\usetikzlibrary{arrows.meta,positioning,backgrounds,calc}
\definecolor{ta}{RGB}{60,90,200}    
\definecolor{tb}{RGB}{230,140,30}   
\definecolor{tc}{RGB}{25,160,150}   
\definecolor{ink}{RGB}{34,40,48}
\definecolor{mut}{RGB}{120,130,140}

\definecolor{lowc}{RGB}{215,48,39}    
\definecolor{midc}{RGB}{255,255,191}  
\definecolor{highc}{RGB}{0,104,55}    
\newcommand{\setcellcolor}[1]{%
  \ifnum#1>50
    \pgfmathtruncatemacro{\pp}{(#1-50)*2}\colorlet{cellc}{highc!\pp!midc}%
  \else
    \pgfmathtruncatemacro{\pp}{#1*2}\colorlet{cellc}{midc!\pp!lowc}%
  \fi
}
\newcommand{\hcell}[3]{%
  \setcellcolor{#3}%
  \ifnum#3>79 \colorlet{txtc}{white}\else\colorlet{txtc}{black!80}\fi
  \fill[cellc, draw=white, line width=1.2pt] (#1*\cw,-#2*\ch) rectangle ++(\cw,-\ch);
  \node[font=\small\bfseries, text=txtc] at (#1*\cw+0.5*\cw,-#2*\ch-0.5*\ch) {#3};
}

\newcommand{\Nano}{VPsy\xspace}
\newcommand{\Flash}{VPsy-Flash\xspace}
\newcommand{\NanoFull}{VisionPsy-Nano-460M\xspace}
\newcommand{\FlashFull}{VisionPsy-Nano-460M-Flash\xspace}
\newcommand{\Fam}{VisionPsy-Nano\xspace}

\usepackage[accsupp]{axessibility}
\usepackage{caption}
\usepackage{fontawesome5}   

\definecolor{pastelgreen}{RGB}{210, 240, 210}
\newcommand{\GT}[1]{\textcolor{green!55!black}{\textbf{[Correct Answer: #1]}}}

\usepackage{fontawesome5}
\newcommand{\hflogo}{\raisebox{-0.15em}{\includegraphics[height=0.95em]{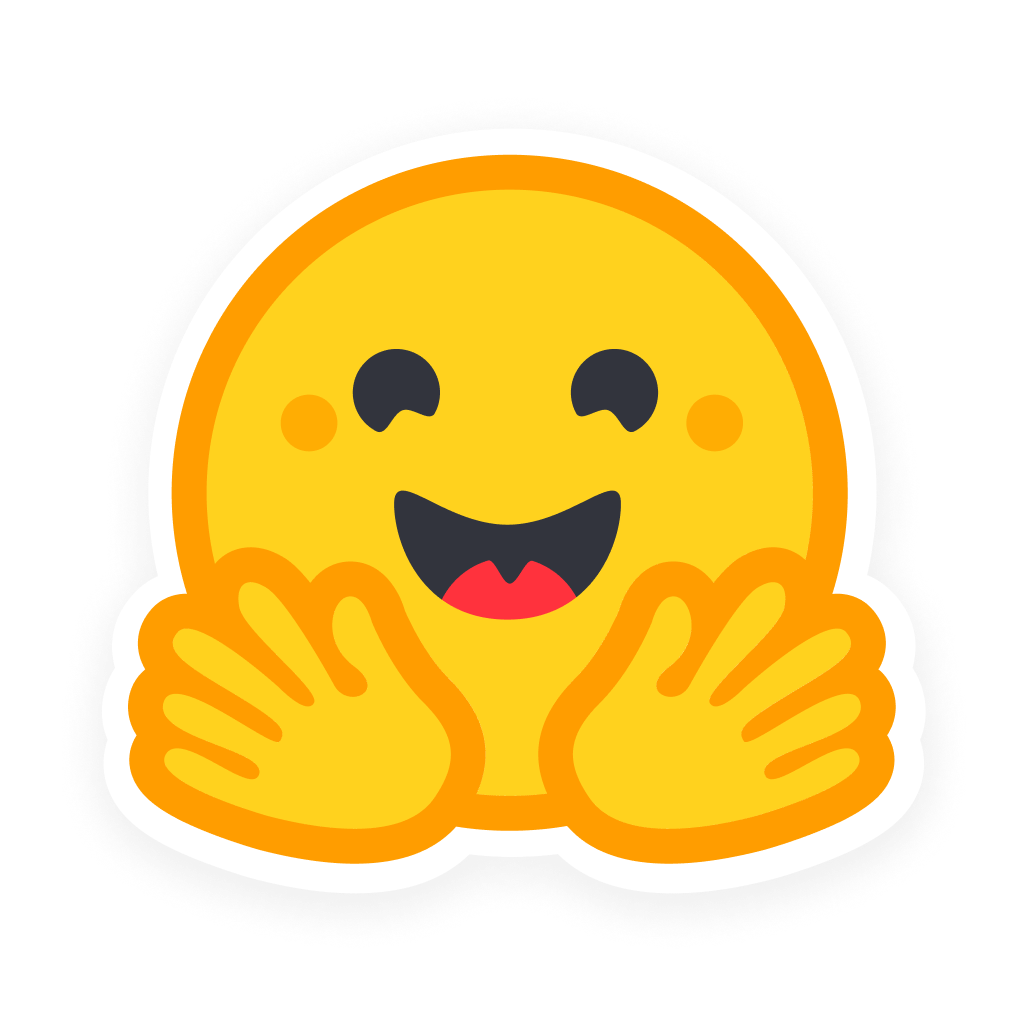}}}
\newcommand{\ghlogo}{\raisebox{-0.05em}{\faGithub}}

\definecolor{wacvblue}{rgb}{0.21,0.49,0.74}
\usepackage[pagebackref,breaklinks,colorlinks,allcolors=wacvblue]{hyperref}

\def\wacvPaperID{883} 
\def\confName{WACV}
\def\confYear{2027}

\title{VisionPsy-Nano: Improving Accuracy, Efficiency, and Reliability in On-Device Vision-Language Models}

\author{
Khurram Azeem Hashmi, \; Mohammadreza Zolfaghari, \; Changdae Park, \; Rishabh Jain, \; Nicholas Moratelli,\\
Pengfei Wei, \; Louis Lu, \; Tianchi Liu, \; and \; Amril Nazir\\[2pt]
Tether AI Research\\[2pt]
{
\href{https://huggingface.co/collections/qvac/visionpsy}{\hflogo\; Models} \quad
\href{https://github.com/tether-ai-research/qvac-visionpsy-nano}{\ghlogo\; Code}
}
}

\begin{document}
\maketitle
\begin{abstract}
Sub-billion-parameter Vision-Language Models are increasingly viable for on-device deployment, yet compact model size alone does not guarantee usability. On a phone, such a model can still require more than two minutes to produce its first token. On-device usability depends on three axes: accuracy, efficiency, and behavioral reliability; standard benchmarks miss the third, with answers too short to expose doom loops and prompts too benign to probe adversarial safety.
We introduce a diagnosis-driven post-training recipe in which a teacher VLM stress-tests the student, uncovers failure modes beyond human priors, and converts them into targeted supervision and preference alignment, supplementing generic data scaling with failure-driven optimization. Coupled with two visual-token policies, the recipe yields two accuracy-efficiency variants with improved behavioral reliability. \textbf{\NanoFull} attains a 62.3 normalized average over 17 benchmarks, the highest among openly released $\sim$0.5B models, +7.4 over its base at identical architecture and token budget, with doom-loop rates at or below the strongest baseline's. \textbf{\FlashFull} retains 61.4 while cutting warm time-to-first-token on a Pixel 9 from 138\,s to 6.1\,s (23$\times$). By jointly addressing all three axes, we move compact VLMs toward practical on-device usability.
\end{abstract}

\section{Introduction}
\label{sec:intro}
\begin{figure}[htbp]
    \centering
    \includegraphics[width=\linewidth]{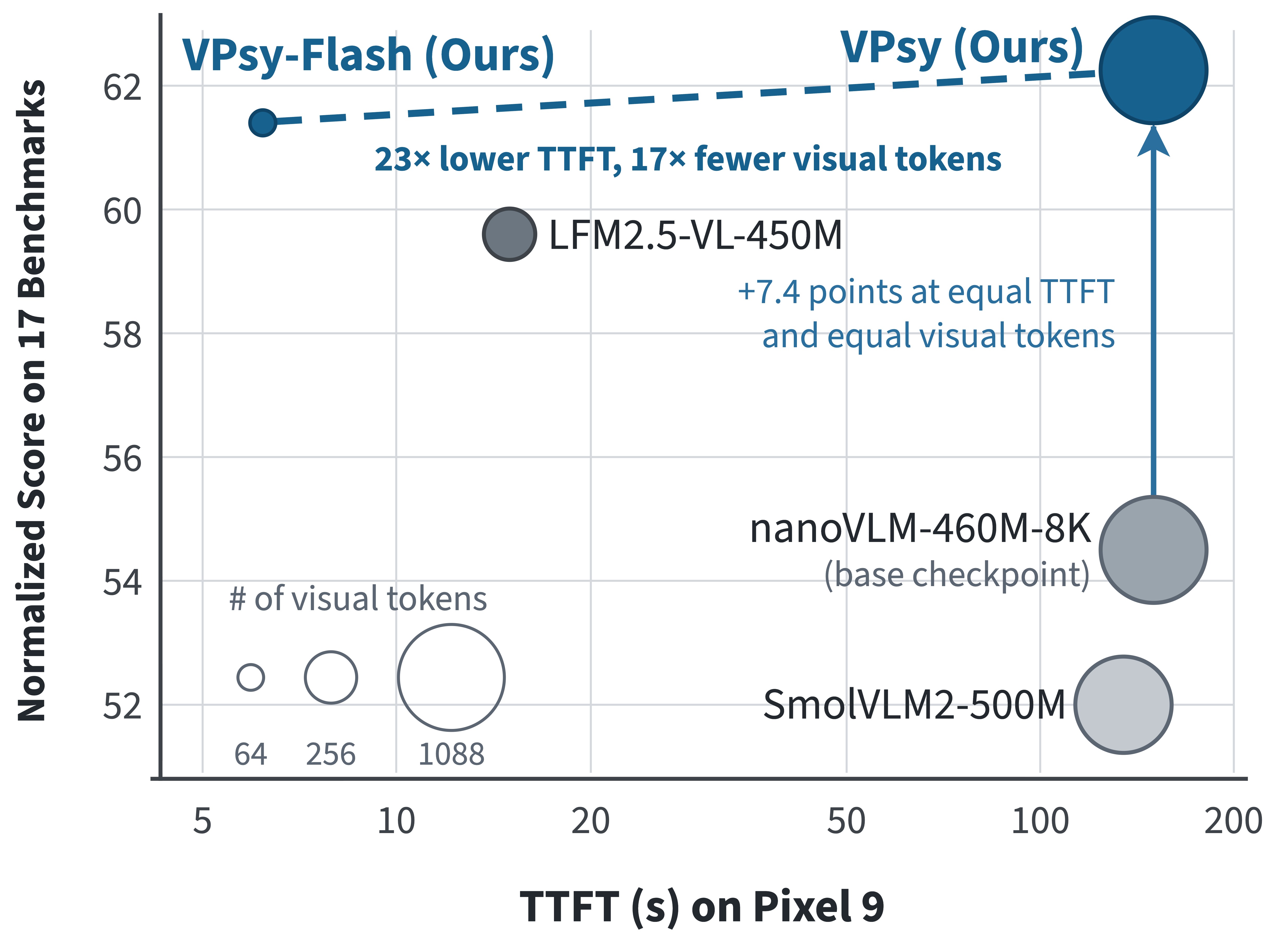}
    \caption{\textbf{Accuracy–latency comparison at the 0.5B scale.} Normalized average over 17 benchmarks against warm TTFT at $512^2$ on a Pixel 9. \Nano\ gains 7.4 points over its nanoVLM-460M-8k starting checkpoint at identical latency and token budget; \Flash\ retains 61.4 of 62.3 at 6.1\,s TTFT. Scores are measured at full precision, TTFT with Q4\_0 weights.} 
    \label{fig:optimal_tradeoff}
\end{figure}

Vision-Language Models (VLMs)~\cite{liu2023llava, team2025gemma, chen2024internvl, yu2026minicpm} are increasingly deployed on smartphones, where local execution keeps content private, works offline, and adds no per-query serving cost. The phone is also a severely constrained place to run a VLM: 6 to 12\,GB of shared memory and tight thermal limits admit only sub-billion-parameter checkpoints under low-bit quantization~\cite{chu2023mobilevlm, liquid2026lfm25vl, marafioti2025smolvlm, qwen2026qwen35}. We study the $\sim$0.5B (450--500M-parameter) tier of this regime, populated by several strong openly released checkpoints~\cite{huggingface2025nanovlm,marafioti2025smolvlm,liquid2026lfm25vl}. Even at this scale, the visual prefix dominates inference cost: at $512^2$ input resolution, 0.5B checkpoints emit 256 to 1{,}088 visual tokens for a single image, and warm time-to-first-token (TTFT) spans seconds to over two minutes on the same phone (\cref{fig:optimal_tradeoff}).

At this scale, choosing which checkpoint to ship is driven almost entirely by benchmark accuracy. Accuracy is the right first axis and carries our headline result, but the shipping decision spans three: accuracy, on-device cost, and behavioral reliability. Neither of the latter two is recoverable from it. Checkpoints within five points of one another span a $9\times$ TTFT range (\cref{fig:optimal_tradeoff}). Short-answer benchmarks terminate after a few tokens, so the degenerate repetition (``doom loops'') of open-ended generation~\cite{doomloop1,doomloop2} develops off-benchmark, and fixed question sets omit adversarial inputs, so safety under text- and image-borne attacks goes unmeasured.

Accuracy at this scale is bounded by what generic supervision can reach. Broad coverage saturates quickly, and the headroom past it is not a data-volume problem but a set of specific, nameable failures that added corpus size leaves intact: supervision targeted at diagnosed failures beats size-matched generic supervision on all four capability areas (\cref{tab:ablation_failure_guided_SFT}). Diagnosis is therefore what carries accuracy past the point where volume stops paying, and it is also the only instrument in the pipeline that observes the behavioral axis and converts it into a training signal. We build the repair stages of the pipeline around it: a teacher VLM~\cite{qwen2026qwen36} stress-tests the student towards the failing regions and maintains a persistent weakness catalogue, and the diagnosis dictates every stage downstream of capability-level supervision.

\begin{figure*}
    \centering
    \includegraphics[width=1\linewidth]{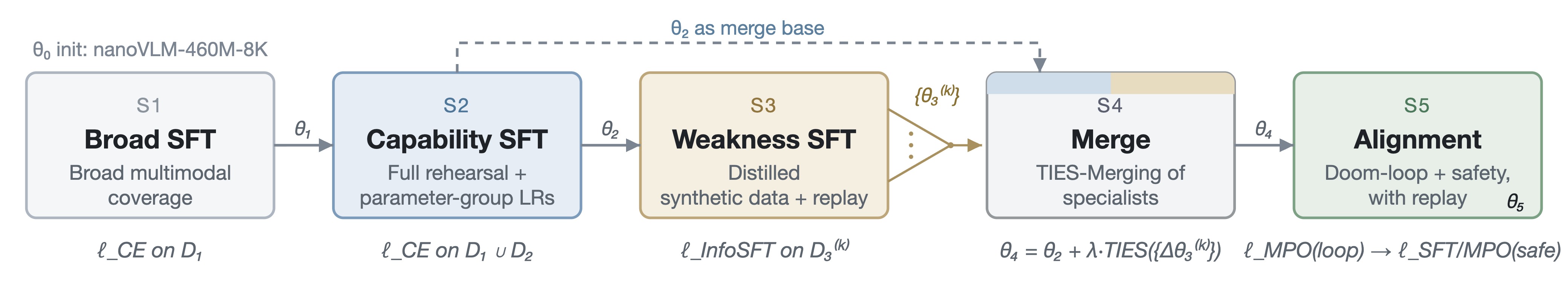}
    \vspace{-20pt}
    \caption{\textbf{The five-stage post-training recipe.} From the
    public nanoVLM-460M-8k checkpoint $\theta_0$, Stage~1 installs
    broad multimodal coverage ($\ell_{\mathrm{CE}}$ on $\mathcal{D}_1$)
    and Stage~2 targets deployment capabilities on
    $\mathcal{D}_1 \cup \mathcal{D}_2$ with full rehearsal, producing
    $\theta_2$, the reference for the failure analysis of
    \cref{sec:failure} and the merge base of Stage~4 (dashed arrow).
    Stage~3 fine-tunes $\theta_2$ into $K$ weakness-targeted
    specialists and Stage~4 TIES-merges~\cite{yadav2023ties} them into one checkpoint
    $\theta_4$ at unchanged architecture and serving cost
    (\cref{fig:mergesynth}). Stage~5 aligns doom-loop termination
    and safety with capability replay, yielding the released
    checkpoint $\theta_5$.}
    \label{fig:pipeline_figure}
    \vspace{-2mm}
\end{figure*}

We instantiate this recipe as five post-training stages on the public nanoVLM-460M-8k checkpoint~\cite{huggingface2025nanovlm}, as shown in~\cref{fig:pipeline_figure}: Stages~1--2 install broad and capability-level coverage with generic supervision. The diagnosis run against that reference dictates the repair stages: weakness-targeted specialists consolidated by merging and preference alignment, with retention enforced as a constraint at every stage after the first. The recipe is post-training only, adds no parameters, and preserves the base serving cost. We run the recipe twice, producing two independently trained checkpoints that differ only in the visual-token policy fixed before Stage~1. \textbf{\NanoFull{} (\Nano)} inherits the default policy, which isotropically upscales small inputs and inflates the visual prefix to 1{,}088 tokens at $512^2$. \textbf{\FlashFull{} (\Flash)} trains every stage under a bounded native-resolution policy that represents the same input with 64 tokens, a $17\times$ smaller prefix (Sec.~\ref{sec:flash}).

We verify these claims by scoring the full cohort under a single harness that measures all three axes at once: 17 benchmarks spanning the four capability areas of the dominant on-device workload, warm TTFT on four smartphones, and long-form and adversarial stress tests. Fig.~\ref{fig:optimal_tradeoff} locates the resulting operating points on the accuracy--latency plane; the reliability axis is reported in Tab.~\ref{tab:doom_loop}, where alignment repairs a doom-loop regression both benchmarks and the aggregate miss, and Fig.~\ref{fig:safety_alignment}, where safety pass rates improve on seven of eight categories.

To summarize, our contributions are:
\begin{itemize}
    \item \textbf{Active failure discovery beyond human priors.} A closed teacher--student loop that stress-tests the student toward its failing regions and rewrites a persistent weakness catalogue, discovering failure categories no human seed covered (\cref{sec:failure}).
    \item \textbf{Diagnosis-driven post-training.} A five-stage recipe in which generic supervision installs coverage and the diagnosis dictates every repair stage beyond it, each constrained to preserve earlier capabilities. Failure-targeted supervision beats size-matched generic supervision on all four capability areas (\cref{tab:ablation_failure_guided_SFT}).
      \item \textbf{Two open, deployable operating points at the $\sim$0.5B scale.} \Nano attains the highest normalized average in the openly released $\sim$0.5B cohort of 62.3, +7.4 over its base at identical architecture and token budget, surpassing LFM2.5-VL-450M~\cite{liquid2026lfm25vl} on 16 of 17 benchmarks (Tab.~\ref{tab:main_sota}). \Flash reaches 61.4 with a 6.1\,s Pixel~9 warm TTFT, against 138\,s at the default policy (23$\times$, Fig.~\ref{fig:optimal_tradeoff}).
\end{itemize}

\section{Related Work}
\label{sec:related}

\noindent\textbf{Edge-efficient vision-language models.}
Work on deployable vision-language models (VLMs) increasingly focuses on both compact backbones and efficient visual tokenization. Qwen3.5-0.8B~\cite{qwen2026qwen35}, MobileVLM~\cite{chu2023mobilevlm}, SmolVLM~\cite{marafioti2025smolvlm}, and LFM2.5-VL-450M~\cite{liquid2026lfm25vl} demonstrate useful multimodal capabilities at mobile-oriented and sub-billion-parameter scales, while FastVLM~\cite{vasu2024fastvlm} highlights that both vision-encoder latency and the number of visual tokens passed to the language model are critical to end-to-end responsiveness. Complementary approaches address visual-token redundancy within the multimodal stack: PACT prunes and clusters visual tokens~\cite{dhouib2025pact}, ATP-LLaVA performs instance- and layer-adaptive pruning~\cite{ye2025atpllava}, and Delta-LLaVA learns a compact visual projector under a constrained token budget~\cite{zamini2026deltallava}. Our work builds on the public nanoVLM implementation~\cite{huggingface2025nanovlm} and targets this complementary efficiency bottleneck.

\noindent\textbf{Failure-guided adaptation and retention.}
Recent VLM research increasingly focuses on adapting models to observed weaknesses: error-driven multimodal tuning retrieves targeted examples based on teacher diagnoses~\cite{yao2024errordriven}, LLaVA-Critic provides multimodal evaluation signals for optimization~\cite{xiong2025llavacritic}, and failure-informed self-augmentation synthesizes training data from model failures~\cite{jiang2026fisa}. 
Note that effective adaptation must also preserve existing capabilities: dual-level adversarial alignment mitigates encoder- and fusion-level domain shifts~\cite{dayal2026duaa}, while pseudo-rehearsal and DREAM address catastrophic forgetting in continual VLM adaptation~\cite{das2025onevlm,chee2026dream}, consistent with the broader benefits of replay~\cite{rolnick2019replay}. 
Building on these directions, our pipeline combines failure mining and teacher-guided data discovery with gold-anchored corrective supervision, while explicitly treating retention as a constraint throughout sequential post-training.

\noindent\textbf{Model merging for capability composition.}
Weight-space composition provides a way to combine complementary specialists without increasing inference cost. Model Soups showed that averaging fine-tuned checkpoints can improve generalization when solutions are compatible~\cite{wortsman2022soups}, and Task Arithmetic represented fine-tuning updates as task vectors relative to a shared base~\cite{ilharco2023taskarithmetic}. When task vectors conflict, however, naive averaging can cancel useful updates. TIES-Merging addresses this issue by sparsifying task vectors, resolving sign disagreement, and averaging only aligned components~\cite{yadav2023ties}. More recently, OptMerge extended model-merging analysis to multimodal LLM capabilities and modalities~\cite{wei2025optmerge}. We use this line of work to consolidate capability-specific specialists trained from the same checkpoint into one model with unchanged architecture and serving cost.

\section{Post-Training Recipe}
\label{sec:method}
We turn a small, permissively licensed VLM~\cite{huggingface2025nanovlm} into a phone-usable checkpoint that is strong on the three axes a shipping decision weighs (\cref{sec:intro}):
\emph{accuracy}, \emph{on-device efficiency}, and \emph{behavioral reliability}, by post-training alone, with no added parameters. Three principles govern the pipeline: \emph{coverage precedes diagnosis and diagnosis precedes repair}, \emph{retention is a hard constraint}, and \emph{consolidation over specialization}. Stages~1 and~2 install broad and capability-level coverage; the diagnosis then runs against this saturated reference and dictates the repair executed by Stages~3--5. The five stages deliver accuracy and behavioral reliability (Fig.~\ref{fig:pipeline_figure}); efficiency follows from the visual-token policy (Sec.~\ref{sec:flash}).

\noindent\textbf{Base checkpoint.}
\label{sec:base}
We use the public 460M-parameter, 8k-context
release of nanoVLM~\cite{huggingface2025nanovlm}, which pairs a SigLIP2
vision encoder~\cite{tschannen2025siglip2} with a SmolLM2 language
backbone~\cite{allal2025smollm2} through a pixel-shuffle MLP connector, as the base model.
We select nanoVLM because it is fully open-source, commercially permissive, and the strongest permissively licensed checkpoint in the ${\sim}0.5$B tier (\cref{tab:main_sota}).\footnote{Although LFM2.5-VL-450M scores higher, its license restricts commercial use above a US\$10M annual-revenue threshold. Hence, its derivatives cannot be released permissively. Both of our checkpoints are released under Apache 2.0.}

\noindent\textbf{Retention as a design constraint.}~\phantomsection\label{sec:retention}
We define retention as the preservation of capabilities acquired in earlier stages, measured as the change in accuracy on capability areas a stage does not target. Every stage after the first narrows the training distribution relative to its predecessor, and at 460M parameters a narrowed distribution degrades untargeted capabilities within a fraction of an epoch. We therefore impose retention as a constraint on every stage after Stage~1, which installs the coverage to retain, and select each mechanism from the stage's objective; the stage descriptions name them, and~\cref{tab:retention-ablation} reports the effect of removing each.

\subsection{Failure Analysis Framework}
\label{sec:failure}
While initial training effectively builds broad foundational capabilities, models often still exhibit residual, specific failure modes that span both capability and behavioral failures, the latter of which do not surface on short-answer benchmarks. Mapping this residual is the role of our failure analysis framework, which combines
passive mining with active discovery to systematically identify and categorize blind spots without relying solely on human priors.
\emph{Passive mining} localizes concrete failures within the labeled data we already hold, returning hard, verified failing items. \emph{Active discovery} looks beyond it: a stronger teacher VLM stress-tests the student and maintains a persistent weakness catalogue whose clusters carry a severity score and the probes that elicit them.

\noindent\textbf{Pipeline Integration.} Within the five-stage pipeline (Fig.~\ref{fig:pipeline_figure}), the framework runs against the checkpoint that Stage~2 produces, so its map characterizes the residual of a model that has already absorbed the broad and capability-targeted supervision. The diagnosis then dictates every repair stage: it selects the Stage-3 repair targets and shapes their synthetic supervision, and it defines the rollout-mining procedure that Stage~5 applies to the merged checkpoint to build its preference pairs (Sec.~\ref{sec:stages}).

\noindent\textbf{Passive mining.} A model-in-the-loop pass has the student answer every item in our training pool, scored automatically by rule-based checks for structured formats and a VLM judge for open-ended answers. We keep only hard, cleanly labeled cases the model gets wrong while the ground truth is reliable, yielding a focused map of concrete weaknesses (knowledge recall, chart arithmetic, spatial reasoning). The same pass defines our rollout-mining procedure for behavioral issues like degenerate repetition (``doom loops''): generating multiple stochastic rollouts per prompt isolates cleanly terminated responses from genuine loops without altering the task or visual input.

\begin{figure}[t]
    \centering
    \includegraphics[width=\linewidth]{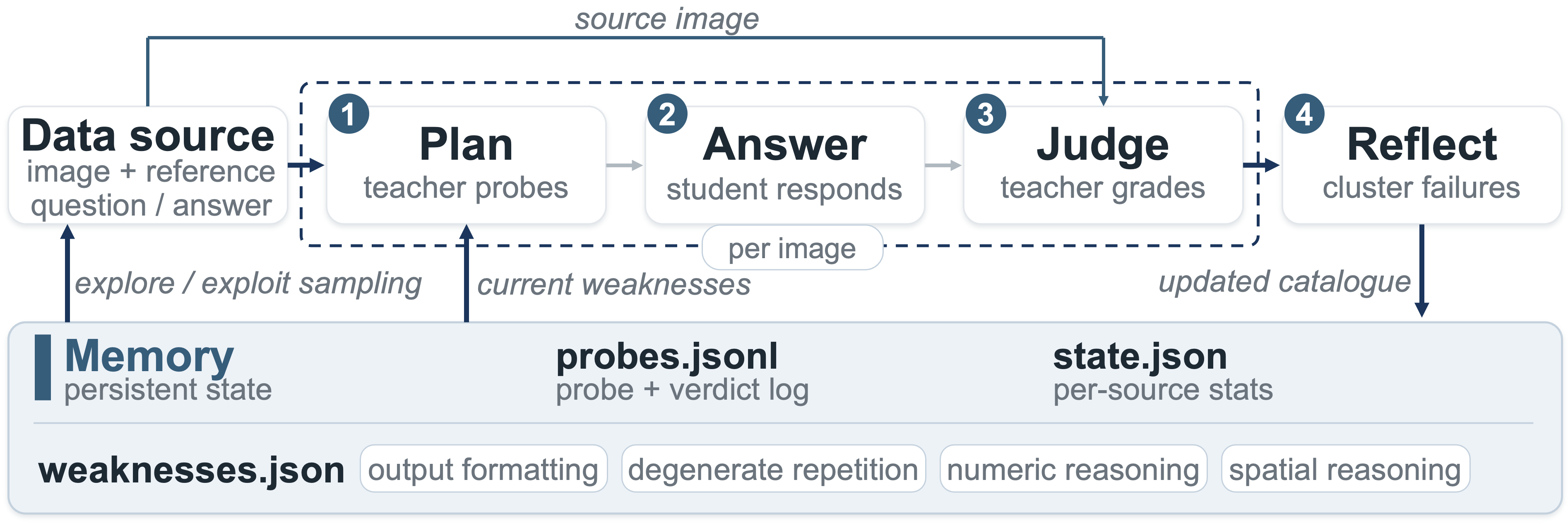}
    \caption{One round of the active failure-discovery loop: a teacher VLM probes the student on images sampled toward its weak spots, grades the answers with the image in view, and reflects over the failures to rewrite the weakness catalogue seeding the next round.}
    \label{fig:active_failure}
\end{figure}

\noindent\textbf{Active discovery.} Passive analysis only maps \emph{existing} data; our active harness instead hunts for blind spots. As shown in Figure~\ref{fig:active_failure}, a strong teacher VLM (Qwen3.6-27B) drives a closed feedback loop. \textbf{Explore/Exploit sampling:} We first sample images from data source $s$ with probability $p_s \propto \epsilon/|S| + (1-\epsilon)\,\hat{r}_s$, where $\hat{r}_s$ is the running failure rate of source $s$ and $\epsilon = 0.35$ is a fixed exploration fraction. This concentrates compute on failing regions while avoiding mode collapse. The loop then proceeds through four stages: 
(1)~\textbf{Plan:} Using the sampled image, reference question and answer, and the current weakness catalogue, the teacher writes $k$ realistic probes that deliberately re-attack known soft spots rather than asking generic questions. 
(2)~\textbf{Answer:} The student answers the $k$ probes through standard inference, ensuring responses match evaluation behavior. 
(3)~\textbf{Judge:} With the source image in view, the teacher grades the $k$ answers jointly, catching character-level slips, empty outputs, and degenerate repetition. 
(4)~\textbf{Reflect:} At the round's end, the teacher clusters a bounded window of fresh failures to rewrite the deduplicated weakness catalogue. Because the probe log and per-source metrics update continuously, sampling and planning always act on current evidence.


\noindent\textbf{Surpassing human priors.} We initialize the weakness catalogue with seeds from a per-capability error analysis of the student, inspection of its failing generations, and failure modes typical of sub-billion VLMs. Over successive rounds the loop rewrote every seeded hypothesis and autonomously discovered failure modes in categories no seed covered, including hallucination, refusals, and generation idiosyncrasies such as gibberish arithmetic. The teacher assigns each cluster a severity, and most novel discoveries reach the maximum. Complete execution statistics and catalogue definitions are in Appendix \textbf{A}.

\subsection{Training Stages}
\label{sec:stages}
\Cref{fig:pipeline_figure} summarizes the five stages. We write $\theta_0$ for the base checkpoint, $\theta_i$ for the checkpoint produced by Stage~$i$,
and $\mathcal{D}_i$ for the corpus introduced at Stage~$i$. \\

\noindent\textbf{Stage 1: Broad SFT.} Stage~1 broadens the multimodal coverage of $\theta_0$ prior to any targeted intervention. Let $\mathcal{D}_1$ denote the Stage-1 corpus of image--question--answer triples from Nemotron-Image-Training-v3~\cite{nvidia2025nemotron}, filtered to single-image samples with chain-of-thought segments stripped (corpus preparation in Appendix~\textbf{D.2}). At 460M parameters the limiting factor is absent competence on entire input categories (dense scene text, tabular documents, chart axes, diagram callouts) rather than reasoning error, so we optimize for breadth of input type: we fine-tune $\theta_0$ on $\mathcal{D}_1$ under token-level cross-entropy $\ell_{\mathrm{CE}}$ over plain answers, yielding $\theta_1$, and defer capability sharpening to subsequent stages (training configuration in Appendix~\textbf{D.2}).\\

\begin{figure}[t]
  \centering
  \includegraphics[width=\linewidth]{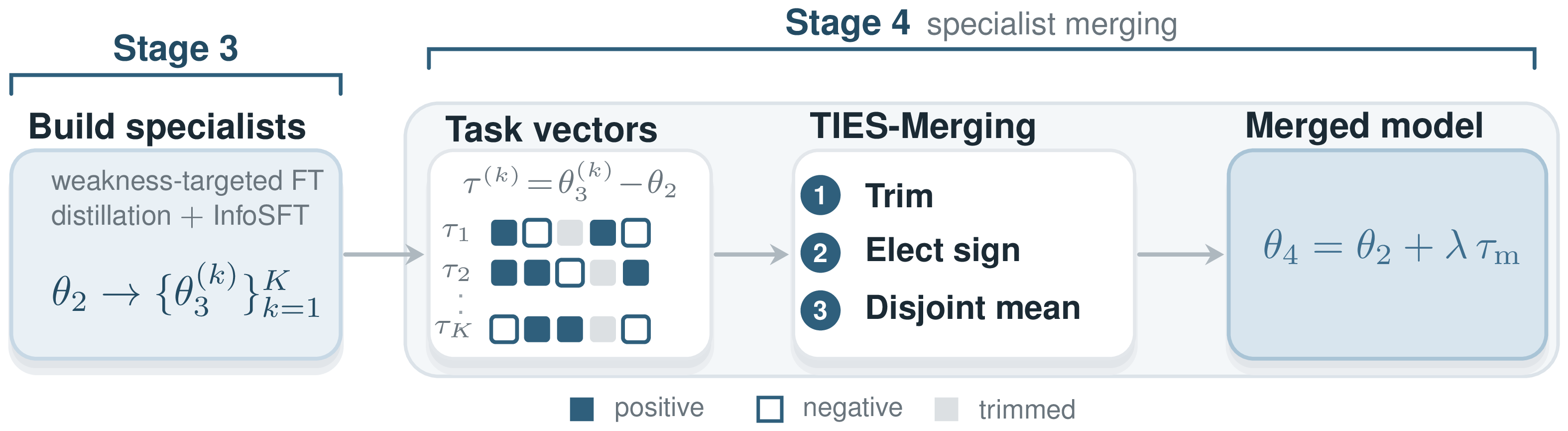}
    \caption{Stages~3--4. Stage~3 turns the base $\theta_2$ into $K$ weakness-targeted
      specialists $\{\theta_3^{(k)}\}$; Stage~4 merges
      their task vectors $\tau^{(k)}=\theta_3^{(k)}-\theta_2$ with TIES into a single
      checkpoint.}
  \label{fig:mergesynth}
  \vspace{-3mm}
\end{figure}

\noindent\textbf{Stage 2: Capability-targeted SFT.}
\label{sec:stage2}
Stage~2 reallocates supervision toward the capabilities the deployment
exercises while holding Stage-1 coverage fixed, training from
$\theta_1$. 

\noindent\textbf{Coverage gaps.} $\mathcal{D}_1$ underrepresents dense OCR, document and chart
understanding, formula recognition, spatial reasoning, and multimodal
instruction following, which together constitute the dominant on-device
workload. We assemble a supplemental corpus $\mathcal{D}_2$ from
multimodal instruction-following data~\cite{mmif23k_iccv25} and a
curated FineVision~\cite{finevision2025} subset covering these axes. All
samples use the Stage-1 $(\text{image},\text{question},\text{answer})$ schema, which keeps output format independent of data
provenance across stages.

\noindent\textbf{Retention control.}
Stage~2 applies two of the mechanisms introduced in \cref{sec:retention}. We train on $\mathcal{D}_1 \cup \mathcal{D}_2$ under $\ell_{\mathrm{CE}}$ rather than on $\mathcal{D}_2$ alone, so Stage-1 coverage is rehearsed while the targeted axes are strengthened. In addition, and only in Stage~2, the encoder, connector, and language backbone form separate parameter groups with distinct learning rates: the largest for the connector, the smallest for the encoder (Stage-2 learning rates are given in Appendix \textbf{D.2}).

The resulting checkpoint $\theta_2$ serves as the reference for the remainder of the pipeline. Both the passive mining pass and the active discovery loop of \cref{sec:failure} run against $\theta_2$, so the weakness map they produce characterizes the residual gaps of a model that has already absorbed the broad and capability-targeted public supervision available to us. These residual failures define the targets
of Stages~3--5. $\theta_2$ is also the base onto which the Stage-3 specialists are merged.

\paragraph{Stage 3: Weakness-targeted fine-tuning.}
\label{sec:stage3}
An analysis of $\theta_2$ still reveals concentrated
weaknesses (knowledge recall, document and infographic reading, chart arithmetic, spatial reasoning). Stage~3 repairs them by finetuning $\theta_2$ into one specialist per diagnosed capability area: grouping the diagnosed weakness clusters of Sec.~\ref{sec:failure} by capability yields $K{=}6$ mixtures $\{\mathcal{D}_3^{(k)}\}_{k=1}^{K}$, so $K$ is set by the diagnosis rather than chosen, and specialists $\{\theta_3^{(k)}\}_{k=1}^{K}$, which Stage~4 merges into one checkpoint (Fig.~\ref{fig:pipeline_figure}). The targets come directly from \cref{sec:failure} with a fixed division of labor: the mined items dictate \emph{what} to repair, supplying the concrete failing cases, and the weakness clusters dictate \emph{how}, naming the data-synthesis recipe for each specialist. 

\noindent\textbf{Synthetic data as distillation.}
Rather than hand-label examples, we distill them: a stronger, vision-enabled
teacher (Qwen3.6-27B~\cite{qwen2026qwen36}) transfers knowledge through the
examples it produces. The pipeline is \emph{gold-anchored}, since each mined item
already carries a trusted answer, we keep it fixed and accept only teacher output
that agrees. For each failure the teacher produces \emph{rationales} (an
image-grounded chain of reasoning to the known answer), \emph{paraphrases} (a
rewording that preserves the answer), or \emph{new questions} (harder questions,
but only where the answer is fully determined by the image content, a value, a
piece of text, or a visible spatial relation, so it can be verified). Every item then clears a verification gate, the teacher re-answers
blind from the image and must agree with itself across repeated attempts and failures are discarded (\cref{fig:mergesynth}).


\noindent\textbf{Objective and mixture.} From two rounds of this generation, plus real weakness-targeted data curated from public sources, each mixture $\mathcal{D}_3^{(k)}$ is filled and carries a replay buffer against forgetting. Unlike the broad corpora of Stages~1--2, these distilled mixtures are dominated by boilerplate tokens, so uniform
cross-entropy would spend its capacity fitting them. We therefore finetune $\theta_2$ separately on each with the InfoSFT objective $\ell_{\mathrm{InfoSFT}}$~\cite{sabbaghi2026infosft}, which down-weights such tokens and concentrates the loss on the informative, medium-confidence ones that carry the targeted capability. The replay buffer is Stage~3's retention mechanism, and the InfoSFT weighting admits the parameter drift that Stage-4 merging exploits.\\

\noindent\textbf{Stage 4: Specialist merging.}
\label{sec:stage4}
Stage~3 yields specialists $\{\theta_3^{(k)}\}_{k=1}^{K}$, each strong in a different area; picking a single ``best overall'' one would trade one capability for another, so Stage~4 fuses them into one checkpoint $\theta_4$ at unchanged architecture and serving cost. We operate in weight space relative to the shared base $\theta_2$: each specialist defines a \emph{task vector} $\tau^{(k)} = \theta_3^{(k)} - \theta_2$, and TIES-Merging~\cite{yadav2023ties} trims, sign-elects, and disjointly averages these vectors (\cref{fig:mergesynth}) so that conflicting updates no longer cancel, yielding $\theta_4 = \theta_2 + \lambda\,\mathrm{TIES}\!\left(\tau^{(1)},\ldots,\tau^{(K)}\right)$. We select each specialist on held-out validation splits and set $\lambda$ to favor the capability areas most in need of improvement; neither is tuned on the benchmarks (\cref{sec:experiments}). $\theta_4$ recovers each specialist's advantage and is stronger overall than any single specialist and than naive averaging, at a small perception trade-off (\cref{tab:merge-ablation}). \\



\begin{table*}[t] 
\centering
\setlength{\tabcolsep}{2.6pt}
\renewcommand{\arraystretch}{1.15}
\resizebox{\textwidth}{!}{%
\begin{tabular}{l ccccc ccccc cccccc cc c}
\toprule
& \multicolumn{5}{c}{\textbf{Document Understanding \& OCR}}
& \multicolumn{4}{c}{\textbf{Visual Perception}}
& \multicolumn{6}{c}{\textbf{Reasoning \& Knowledge}}
& \multicolumn{2}{c}{\begin{tabular}{@{}c@{}}\textbf{IF \& Reliab.}\end{tabular}} 
& \\
\cmidrule(lr){2-6}\cmidrule(lr){7-10}\cmidrule(lr){11-16}\cmidrule(lr){17-18}
\textbf{Model}
& OCRB$^{\dagger}$ & DocVQA$^{\dagger}$ & ChartQA$^{\dagger}$ & InfoVQA$^{\dagger}$ & TextVQA$^{\dagger}$
& MME & SEED & MMB & RWQA
& SQA & AI2D & MMStar & MMMU & MathV & MMVet
& \makebox[3.5em][c]{MM-IF} & \makebox[3.5em][c]{POPE} 
& \textbf{Avg$_{\text{norm}}$} \\
\midrule
SmolVLM2-500M~\cite{marafioti2025smolvlm}
& 619 & 74.7 & 64.9 & 39.0 & 71.3
& 1455 & 62.1 & 51.4 & 50.1
& 76.3 & 57.3 & 38.3 & \textbf{31.6} & 38.5 & 28.6
& 11.1 & 82.7 
& 52.5 \\
nanoVLM-460M-8k~\cite{huggingface2025nanovlm}
& 745 & 82.3 & 70.4 & 46.0 & 74.5
& 1442 & 64.3 & 53.1 & 51.8
& 76.8 & 55.3 & 37.0 & 31.3 & 33.3 & 28.4
& 20.2 & 82.7  
& 54.9 \\
LFM2.5-VL-450M~\cite{liquid2026lfm25vl}
& 710 & 82.7 & 76.6 & 48.9 & 78.8
& 1453 & 67.5 & 56.8 & 59.0
& 77.7 & 62.2 & 42.7 & 30.7 & 42.2 & \textbf{36.0}
& 42.0 & 86.5 
& 59.6 \\
\midrule
\textbf{VisionPsy-Nano-460M}
& \textbf{757} & \textbf{85.7} & \textbf{78.7} & 49.8 & \textbf{79.3}
& 1541 & \textbf{69.2} & \textbf{61.9} & \textbf{60.0}
& \textbf{86.5} & \textbf{66.5} & \textbf{47.6} & 31.4 & \textbf{48.9} & 32.3
& 42.3 & \textbf{87.9} 
& \textbf{62.3} \\
\textbf{VisionPsy-Nano-460M-Flash}
& 738 & 85.1 & 77.4 & \textbf{51.1} & 75.6
& \textbf{1619} & 67.6 & 60.5 & 58.4
& 84.0 & 66.4 & 45.5 & 30.7 & 47.8 & 31.1
& \textbf{43.7} & 87.5 
& 61.4 \\
\bottomrule
\end{tabular}%
}
\caption{\textbf{Per-benchmark comparison across the $\sim$0.5B cohort on all 17 public benchmarks.} $^{\dagger}$Scores judged with Qwen3.6-27B~\cite{qwen2026qwen36}.}
\label{tab:main_sota}
\end{table*}

\noindent\textbf{Stage 5: Preference alignment.}
\label{sec:stage5}
As illustrated in Fig.~\ref{fig:pipeline_figure}, Stage~5 addresses two behavioral failures identified by our failure analysis~(\cref{sec:failure}): degenerate repetitive generation (\emph{doom loops}) and unsafe behavior, using preference
pairs mined from stochastic rollouts of the merged checkpoint
$\theta_4$.

\noindent\textbf{Doom-loop correction.} Doom loops concentrate in longer, open-ended generations, where repetition becomes self-reinforcing and prevents termination~\cite{doomloop1,doomloop2}. Because the same image--prompt pair can yield either a clean completion or a loop across stochastic rollouts, the task capability is typically present while the generation behavior is unstable; we therefore cast correction as preference learning, pairing a cleanly terminated rollout with a detected loop for each input. Candidate loops are identified from repeated $n$-grams, lines, and sentences, filtered to exclude ordinary verbosity or truncation. We optimize $\theta_4$ on these self-mined pairs with MPO~\cite{wang2024mpo} together with capability replay, stabilizing response completion before safety alignment.

\noindent\textbf{Safety alignment.} We then align the model against harmful and adversarial requests using a curated subset of public multimodal safety data~\cite{safety_data} plus additional curated cases spanning harmful requests, privacy, self-harm, fraud, and visual prompt injection. We first run safety SFT on curated target responses, then MPO on model-specific preference pairs chosen from multiple rollouts by safety, helpfulness, coherence, and refusal quality. Since aggressive safety optimization can increase over-refusal on benign inputs, we use a smaller learning rate and fewer steps while including benign boundary cases and capability replay throughout. This improves safety behavior while limiting drift from the multimodal capabilities acquired earlier.

\section{\Flash: Training Under a Reduced Visual-Token Budget}
\label{sec:flash}
To mitigate the computational bottlenecks hindering on-device deployment, we introduce \Flash: an efficiency-oriented model that executes the full five-stage recipe of Sec.~\ref{sec:method} under a substantially reduced visual-token budget. Crucially, \Flash requires no architectural changes, additional parameters, or runtime token-compression heuristics. The two runs diverge only in the visual-token policy fixed before Stage~1: \Nano inherits the default full-resolution preprocessing, while \Flash trains every stage under our bounded native-resolution policy. Since zero-shot token restriction degrades accuracy, most sharply on document understanding tasks (Tab. \textbf{17} in the Appendix), executing the recipe under the budget is what preserves it.

\noindent\textbf{Visual-token budget as the binding cost.}
The \Nano visual pipeline represents an image via one global view and $N_{\mathrm{local}}$ local tiles. Since each view yields exactly 64 projected embeddings, the total visual-prefix length (excluding special tokens) is $N_{\mathrm{vis}} = 64 \cdot (1 + N_{\mathrm{local}})$. For images fitting a single $512^2$ tile ($N_{\mathrm{local}} = 0$), the global view suffices; local tiling is bypassed, yielding a minimal prefix of $N_{\mathrm{vis}} = 64$. As these tokens enter the autoregressive context, $N_{\mathrm{vis}}$ directly dictates the prefill latency and memory footprint.

\noindent\textbf{Bounded native-resolution preprocessing.}
The default visual-token policy, inherited by \Nano, isotropically resizes inputs to a 2048-pixel maximum prior to tiling. For smaller real-world inputs, this indiscriminately upscales the image, generating redundant tiles that artificially inflate the token budget without adding visual content. To maximize information density, \Flash enforces a three-stage dynamic policy: (i)~\textbf{Floor scaling:} short sides below 512 pixels are upscaled to 512 to preserve spatial detail; (ii)~\textbf{Ceiling scaling:} long sides exceeding 2048 are downscaled to bound computational overhead; and (iii)~\textbf{Exact-fit snapping:} intermediate images are minimally resized to exact multiples of the 512-pixel tile size. If $H = W = 512$, $N_{\mathrm{local}} = 0$; otherwise, $N_{\mathrm{local}} = (H/512) \times (W/512)$. This ensures the Vision Transformer operates strictly at native tile granularity without computational padding masks.

\section{Experiments}
\label{sec:experiments}

\begin{table*}[t]
\centering
\setlength{\tabcolsep}{2.6pt}
\renewcommand{\arraystretch}{1.15}
\resizebox{\textwidth}{!}{%
\begin{tabular}{l ccccc cccc cccccc cc c}
\toprule
& \multicolumn{5}{c}{\textbf{Document Understanding \& OCR}}
& \multicolumn{4}{c}{\textbf{Visual Perception}}
& \multicolumn{6}{c}{\textbf{Reasoning \& Knowledge}}
& \multicolumn{2}{c}{\begin{tabular}{@{}c@{}}\textbf{IF \& Reliab.}\end{tabular}}
& \\
\cmidrule(lr){2-6}\cmidrule(lr){7-10}\cmidrule(lr){11-16}\cmidrule(lr){17-18}
\textbf{Stage}
& OCRB$^{\dagger}$ & DocVQA$^{\dagger}$ & ChartQA$^{\dagger}$ & InfoVQA$^{\dagger}$ & TextVQA$^{\dagger}$
& MME & SEED & MMB & RWQA
& SQA & AI2D & MMStar & MMMU & MathV & MMVet
& \makebox[3.5em][c]{MM-IF} & \makebox[3.5em][c]{POPE}
& \textbf{Avg$_{\text{norm}}$}  \\
\midrule
$\theta_0$: nanoVLM-460M-8k~\cite{huggingface2025nanovlm}
& 745 & 82.3 & 70.4 & 46.0 & 74.5
& 1442 & 64.3 & 53.1 & 51.8
& 76.8 & 55.3 & 37.0 & 31.3 & 33.3 & 28.4
& 20.2 & 82.7
& 54.9  \\
\midrule
$\theta_1$: + S1 Broad SFT
& 714 & 85.1 & 74.7 & 47.3 & 78.1
& \textbf{1555} & 69.1 & \textbf{62.8} & 56.1
& 85.9 & \textbf{67.8} & 44.6 & \textbf{34.1} & \textbf{50.2} & \textbf{33.7}
& 17.0 & 86.0
& 60.0 \\
$\theta_2$: + S2 Capability-targeted SFT
& 727 & 83.7 & 75.5 & 44.5 & 78.0
& 1545 & 69.4 & 61.0 & \textbf{61.2}
& 86.2 & 66.0 & 47.8 & 32.9 & 45.3 & 32.5
& 42.9 & 87.6
& 61.3 \\
$\theta_3\!\rightarrow\!\theta_4$: + $\{\text{S3}\}\!\rightarrow\!\text{S4}$ Specialist merging
& 753 & 85.5 & 77.8 & 49.2 & \textbf{79.4}
& 1542 & \textbf{69.8} & 61.8 & 59.2
& \textbf{86.6} & 67.3 & \textbf{48.3} & 31.8 & 48.2 & 32.6
& \textbf{43.4} & 87.5
& \textbf{62.3}  \\
\midrule
$\theta_5$: + S5 Preference alignment (MPO)
& \textbf{757} & \textbf{85.7} & \textbf{78.7} & \textbf{49.8} & 79.3
& 1541 & 69.2 & 61.9 & 60.0
& 86.5 & 66.5 & 47.6 & 31.4 & 48.9 & 32.3
& 42.3 & \textbf{87.9} & \textbf{62.3} \\
\bottomrule
\end{tabular}%
}
\caption{\textbf{Stage-wise ablation.} Stage~5 targets behavioral reliability (\cref{tab:doom_loop}, \cref{fig:safety_alignment}) at a flat aggregate. Best scores are \textbf{highlighted}.}
\label{tab:stage_ablation}
\end{table*}

\subsection{Accuracy}
\noindent\textbf{Evaluation Benchmarks.}
We evaluate on \num{17} public benchmarks spanning four capability areas:
document understanding and OCR (OCRBench~\cite{liu2024ocrbench},
DocVQA~\cite{mathew2021docvqa}, ChartQA~\cite{masry2022chartqa},
InfoVQA~\cite{mathew2022infographicvqa}, TextVQA~\cite{singh2019textvqa}), visual
perception (MME~\cite{fu2026mme}, SEEDBench~\cite{li2023seedbench},
MMBench~\cite{liu2024mmbench}, RealWorldQA~\cite{xai2024realworldqa}), reasoning
and knowledge (ScienceQA~\cite{lu2022scienceqa}, AI2D~\cite{kembhavi2016ai2d},
MMStar~\cite{chen2024mmstar}, MMMU~\cite{yue2024mmmu},
MathVista~\cite{lu2024mathvista}, MMVet~\cite{yu2023mmvet}), and instruction
following and reliability (MM-IFEval~\cite{mmif23k_iccv25},
POPE~\cite{li2023pope}). All models are scored under a single
VLMEvalKit~\cite{duan2024vlmevalkit} harness with each benchmark's official
metric, so every entry is directly comparable and reproducible. 
The five free-form benchmarks marked $\dagger$ are scored
with an LLM judge (Qwen3.6-27B~\cite{qwen2026qwen36}\footnote{All experiments use the FP8 variant.}), since exact string match
conflates reading accuracy with output-format compliance. Re-scoring under three LLM judges, two of them (GPT-4o-mini, GPT-5-chat) fully independent of our training pipeline, leaves the ordering and the aggregate margin unchanged. Refer to Appendix~\textbf{D.1} for more details.

\noindent\textbf{Compared Models.}
We compare against the strongest publicly released $\sim$0.5B (450--500M-parameter) vision-language
models: nanoVLM-460M-8k~\cite{huggingface2025nanovlm},
SmolVLM2-500M~\cite{marafioti2025smolvlm} and
LFM2.5-VL-450M~\cite{liquid2026lfm25vl}. nanoVLM-460M-8k is also the base
checkpoint $\theta_0$ of our pipeline, so comparing against it isolates the contribution of post-training from that of the architecture. 

\noindent\textbf{Comparison at the 0.5B scale.} Tab.~\ref{tab:main_sota} reports the cohort under a single VLMEvalKit~\cite{duan2024vlmevalkit} harness. \Nano attains the highest Avg$_{\text{norm}}$, outperforming LFM2.5-VL-450M on 16 of 17 benchmarks; our two checkpoints hold the best cohort score on 15 of 17, conceding only MMVet to LFM2.5-VL and MMMU to SmolVLM2-500M. The 7.4-point gain over the nanoVLM-460M-8k base comes at identical architecture, latency, and token budget, isolating post-training. Under its bounded visual-token policy, \Flash trails \Nano by 0.9 points while posting the cohort-best MME, InfoVQA, and MM-IFEval. The result is robust to the evaluator: all three LLM judges preserve the 16-of-17 margin over LFM2.5-VL and a 2.7--2.9-point aggregate gap (see Appendix~\textbf{C.1}).

\subsubsection{Ablation Study}
\label{sec:ablation_study}

\noindent\textbf{Each stage plays a distinct role.} \Cref{tab:stage_ablation} evaluates the checkpoint after each stage on the full \num{17}-benchmark suite. Broad SFT (S1) accounts for most of the aggregate gain, raising Avg$_{\text{norm}}$ from 54.9 to 60.0 by installing general multimodal coverage. Capability-targeted SFT (S2) adds $+1.3$ to 61.3, concentrated on instruction following and reliability (MM-IFEval $17.0\!\to\!42.9$, POPE $86.0\!\to\!87.6$). From there the aggregate saturates and the later stages act on axes the mean obscures. Stages 3--4 fine-tune $\theta_2$ into weakness-targeted specialists and merge them into $\theta_4$, adding $+1.0$ to 62.3; the diagnosis-driven checkpoints $\theta_4$ and $\theta_5$ hold the best score on 10 of 17 benchmarks, and the merged checkpoint is stronger than any single specialist (\cref{tab:merge-ablation}). Alignment (Stage 5) holds the aggregate flat at 62.3 while cutting the doom-loop rate from 4.30\% to 0.25\% (\cref{tab:doom_loop}), so its contribution is behavioral. No stage produces a net aggregate regression.

\noindent\textbf{Failure-guided SFT vs.\ generic scaling.}
\Cref{tab:ablation_failure_guided_SFT} isolates the value of failure
guidance under matched selection. From the same $\theta_2$, we
fine-tune specialists on size-matched random pools and on the
failure-guided mixtures, each best-selected on held-out splits. The
best random specialist leaves every axis flat or slightly degraded,
while the best failure-guided specialist improves document
understanding and holds the rest: generic supervision
re-fits what the model already does, and only supervision aimed at
the diagnosed residual converts gradient budget into new capability.

\begin{table}
  \centering
  \small
  \setlength{\tabcolsep}{4.5pt}

\begin{subtable}{\linewidth}
\centering
\resizebox{.9\linewidth}{!}{%
\begin{tabular}{lcccc}
\toprule
Method & Doc/OCR & Percep. & Reason. & Instr. \\
\midrule
$\theta_2$ (Stage-2 base)                        & 70.9 & 61.7 & 51.8 & \textbf{65.3} \\
Best random specialist                      & 70.3 & 61.8 & 51.5 & 64.5 \\
Best Failure guided ($\theta_3^{(k)}$)     & \textbf{72.5} & \textbf{61.9} & \textbf{51.9} & 65.2 \\
\bottomrule
\end{tabular}
}
\caption{Failure guided vs.\ random specialist supervision from $\theta_2$.}
\label{tab:ablation_failure_guided_SFT}
\end{subtable}

  \begin{subtable}{\linewidth}
    \centering
    \resizebox{.9\linewidth}{!}{%
    \begin{tabular}{lcccc}
      \toprule
      Method & Doc/OCR & Percep. & Reason. & Instr. \\
      \midrule
      Best single specialist     & 72.5 & \textbf{61.9} & 51.9 & 65.2 \\
      Naive averaging            & 72.1 & 61.8 & 52.2 & 65.1 \\
      TIES-Merging~\cite{yadav2023ties} & \textbf{73.4} & 61.5 & \textbf{52.5} & \textbf{65.4} \\
      \bottomrule
    \end{tabular}
    }
    \caption{Merging: TIES vs.\ averaging vs.\ best specialist.}
    \label{tab:merge-ablation}
  \end{subtable}

  \begin{subtable}{\linewidth}
    \centering
    \resizebox{.9\linewidth}{!}{%
    \begin{tabular}{lcccc}
      \toprule
      Method & Doc/OCR & Percep. & Reason. & Instr. \\
      \midrule
      $\theta_1$ (Stage-2 init)  & \textbf{71.3} & 60.9 & \textbf{52.7} & 51.5 \\
      Single: $\theta_0$ on $\mathcal{D}_1{\cup}\mathcal{D}_2$ & 65.3 & 56.9 & 49.8 & \textbf{66.2} \\
      $\theta_2$ w/o retention   & 61.8 & 54.1 & 47.5 & 58.8 \\
      $\theta_2$ (full Stage 2)  & 70.9 & \textbf{61.7} & 51.8 & 65.3 \\
      \bottomrule
    \end{tabular}
    }
    \caption{Stage-2 retention and staging.}
    \label{tab:retention-ablation}
  \end{subtable}
  
    \caption{\textbf{Accuracy-axis ablations of the training stages.} \subref{tab:ablation_failure_guided_SFT}~The best failure-guided specialist improves $\theta_2$ where the best size-matched random-pool specialist does not. \subref{tab:merge-ablation}~TIES~\cite{yadav2023ties} leads naive averaging and the best single specialist on three of four axes at unchanged inference cost. \subref{tab:retention-ablation}~Stage-2 retention and staging: training on $\mathcal{D}_2$ alone degrades every untargeted axis; full Stage~2 recovers them; single-stage $\mathcal{D}_1{\cup}\mathcal{D}_2$ trails staged $\theta_2$.}
  \label{tab:capability_ablations}
\end{table}

\noindent\textbf{Why merge, and why TIES.} Tab.~\ref{tab:merge-ablation} validates both Stage-4 choices with the same $\theta_2$-initialized specialists. No specialist leads every axis, and naive averaging already beats the best single one on reasoning: the specialists are complementary, so consolidation preserves strengths that selection alone would forfeit, at unchanged inference cost. Averaging, however, cancels updates where task vectors disagree in sign. TIES~\cite{yadav2023ties} resolves these conflicts and leads on three of four axes, trading a small perception dip.

\begin{table}
    \centering
    \small
    \setlength{\tabcolsep}{2.5pt}
    \resizebox{.8\linewidth}{!}{%
    \begin{tabular}{@{}lcccc@{}}
        \toprule
        Model
        & Pixel 9
        & S23
        & S25 Ultra
        & iPhone 15 \\
        \midrule
        nanoVLM (\Nano)
        & 138.3
        & 116.7
        & 58.8
        & 10.9 \\
        LFM2.5-VL
        & 14.9
        & 13.6
        & 5.7
        & 0.4 \\
        SmolVLM2
        & 128.2
        & 113.7
        & 48.7
        & 10.7 \\
        \textbf{\Flash}
        & \textbf{6.1}
        & \textbf{5.9}
        & \textbf{2.6}
        & \textbf{0.3} \\
        \bottomrule
    \end{tabular}
    }
    \caption{\textbf{Warm TTFT (s) at $512^2$ on smartphones.} \textit{\Flash} is the fastest on every device (23$\times$ over the base on Pixel 9). Since \Nano\ shares nanoVLM's architecture and token budget, the base row gives its TTFT. Best lane per model-device pair.}
    \label{tab:flash_ttft}
\end{table}

\begin{table}
\centering
\small
\resizebox{.7\linewidth}{!}{%
\begin{tabular}{lcc}
\toprule
Model & Doom loops & Rate $\downarrow$ \\
\midrule
nanoVLM-460M-8k ($\theta_0$)            & 4 / 395 & 1.01\% \\
SmolVLM2-500M      & 6 / 395 & 1.52\% \\
LFM2.5-VL-450M     & 2 / 395 & 0.51\% \\
\midrule
\Nano ($\theta_{4}$ )       & 17 / 395 & 4.30\% \\
\Nano ($\theta_{5}$ )     & \textbf{1 / 395} & \textbf{0.25\%} \\
\Flash ($\theta_{4}$ )       & 6 / 395 & 1.52\% \\
\Flash  ($\theta_{5}$ )     & \textbf{2 / 395} & \textbf{0.51\%} \\
\bottomrule
\end{tabular}}
\caption{\textbf{Doom-loop rate on the 395-prompt long-form stress test.} Post-training amplifies the failure in both runs ($\theta_0\!\to\!\theta_4$). Stage-5 alignment repairs both to the level of the strongest baseline or below. Lower is better.}
\label{tab:doom_loop}
\end{table}

\noindent\textbf{Why Retention and why separate stage 2.} \Cref{tab:retention-ablation} isolates the Stage-2 design. Fine-tuning $\theta_1$ on $\mathcal{D}_2$ alone gains the targeted instruction axis and degrades every untargeted one, exactly the narrowed-distribution failure Sec.~\ref{sec:method} predicts; with $\mathcal{D}_1$ rehearsal and parameter-group learning rates, $\theta_2$ gains more on the targeted axis while holding all others within one point of $\theta_1$. Single-stage $\mathcal{D}_1{\cup}\mathcal{D}_2$ trails staged $\theta_2$ on every untargeted axis; staging also yields $\theta_1$ as retention reference and $\theta_2$ as a fixed pre-diagnosis checkpoint.
 
\subsection{On-Device Efficiency}
\label{sec:on_device}

We evaluate whether the reduced visual-token budget of \Flash translates into strictly lower inference costs on consumer hardware. We measure the warm TTFT across four smartphones: Google Pixel 9, Samsung Galaxy S23, Samsung Galaxy S25 Ultra, and Apple iPhone 15. All models are benchmarked using a custom \texttt{llama.cpp} backend with Q4\_0 language and Q8\_0 vision weights. Accuracy is reported at full precision throughout, following the standard convention of edge-model evaluations~\cite{liquid2025lfm2}, which report accuracy at full precision and on-device cost under deployed quantization. Q4\_0 is the one format every cohort architecture supports without calibration, so latency is compared on equal footing; per-format accuracy retention is characterized in Tab.~\textbf{9} in the Appendix.

\begin{figure}[t]
    \centering
    \includegraphics[width=\linewidth]{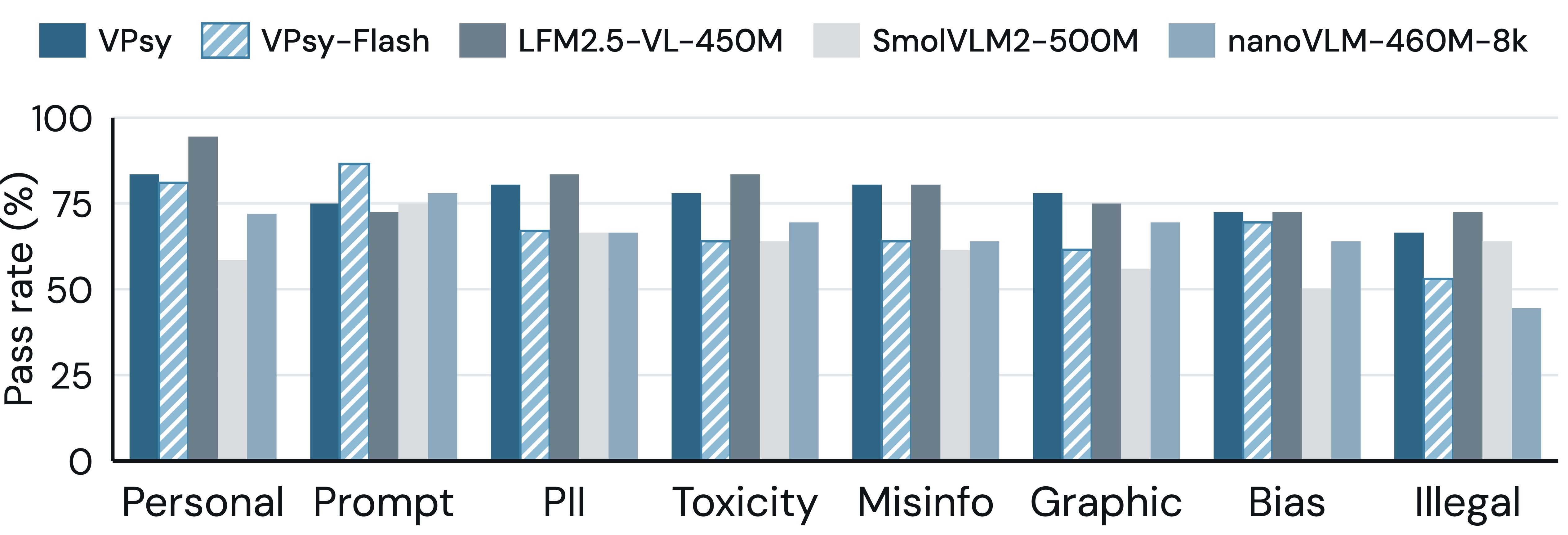}
    \caption{Safety pass rates on eight vulnerability categories under the 288-case red-teaming protocol; higher is better.}
    \label{fig:safety_alignment}
    \vspace{-3mm}
\end{figure}

\noindent\textbf{Time-to-first-token.} \cref{tab:flash_ttft} reports warm TTFT at $512^2$, the resolution where \Flash eliminates the upsampling penalty. \Flash posts the lowest TTFT on every device, 19.8$\times$ to 36.3$\times$ faster than nanoVLM; on the Pixel~9 this turns an unusable 138.3\,s wait into an interactive 6.1\,s, a 23$\times$ speedup that stems directly from the 17$\times$ smaller token prefix (\cref{fig:optimal_tradeoff}). \Flash dominates the strongest cohort model on both axes at once: it leads LFM2.5-VL-450M on 13 of 17 benchmarks (+1.8 Avg$_{\text{norm}}$, \cref{tab:main_sota}) while running
1.3$\times$ to 2.4$\times$ faster on every phone tested.\footnote{Absolute iOS--Android latency gaps largely reflect
llama.cpp backend maturity (Metal vs.\ OpenCL/Vulkan); see
Appendix~\textbf{B.4}.}

\subsection{Behavioral Reliability}
\label{sec:behavior}

Beyond accuracy and efficiency, we evaluate the behavioral reliability targeted by Stage~5: termination under long-form generation and safety under adversarial requests.

\noindent\textbf{Degenerate repetition.} We evaluate doom loops on a stress set of 395 prompts drawn from public multimodal datasets~\cite{vero600k} and manually curated cases, spanning procedural questions, long-form STEM and chart reasoning, exhaustive extraction, and structured outputs. A response counts as a doom loop when it enters persistent phrase-, sentence-, line-, or formatting-level repetition without recovering, detected with the rule-based procedure of Stage-5 mining (repeated $n$-grams, lines, and sentences; verbosity and truncation excluded) applied uniformly to all models. The stress test exposes a failure the 17 accuracy benchmarks are structurally unable to observe: supervised post-training amplifies doom looping in both independent runs, 1.01\% at $\theta_0$ to 4.3\% and 1.52\% at the respective $\theta_4$, while the aggregate improves monotonically (\cref{tab:stage_ablation}). The diagnosis surfaces this regression and supplies its repair from the same checkpoint: Stage~5 optimizes preference pairs mined from $\theta_4$ rollouts (Sec.~\ref{sec:failure}), repairing both runs to 0.25\% and 0.51\% (\cref{tab:doom_loop}). Long-form reliability therefore requires its own instrument and its own training signal.

\noindent\textbf{Adversarial safety.} We evaluate safety with a VLM-native red-teaming suite of 288 adversarial cases across eight vulnerability categories and nine attack strategies, spanning visual attacks with harmful instructions embedded in images and text-based jailbreaks. Harmful intents are hand-curated and every adversarial image is rendered procedurally with attack recipes adapted from OpenRT~\cite{OpenRT2026}. All models face the same fixed attack bank, judged by Qwen3.6-27B~\cite{qwen2026qwen36}. A response passes when the model safely refuses or redirects the request. \Nano improves over nanoVLM on seven of eight vulnerability categories (\cref{fig:safety_alignment}), with the largest gains on illegal activity ($44\%\!\rightarrow\!67\%$), PII leakage ($67\%\!\rightarrow\!81\%$), and misinformation ($64\%\!\rightarrow\!81\%$). \Flash (\cref{fig:safety_alignment}) leads all five models on prompt leakage (86\%) and remains above its base overall (68.1\%).

\section{Limitations}
\label{sec:limitations} Our study is deliberately scoped to the 0.5B, on-device regime and a single base architecture; whether the recipe transfers to larger models or other backbones is left to future work. The active failure-discovery loop is bounded by the teacher: failure modes the teacher cannot itself detect remain outside the weakness catalogue. Our evaluation is automatic over 17 public benchmarks; we mitigate metric and judge sensitivity by confirming that our conclusions hold across three judges. Finally, the safety assessment covers a fixed adversarial suite of eight vulnerability categories and nine attack strategies and should be read as a controlled probe rather than a guarantee of robustness in the wild. Transfer of the recipe to larger scales and other backbones, and teacher ensembles that widen the detectable failure surface, are the immediate extensions.

\section{Conclusion}
\label{sec:conclusion}


We introduce \Nano{} and \Flash{}, two independently trained, deployable checkpoints at the ${\sim}0.5$B tier. \Nano{} attains the highest normalized average (62.3) across 17 benchmarks among openly released ${\sim}0.5$B models. \Flash{} targets the visual prefix bottleneck, reaching 61.4 at a 6.1\,s Pixel~9 warm TTFT, $23{\times}$ below the default policy. Both result from a diagnosis-driven post-training recipe whose active teacher--student loop uncovers and repairs failure modes beyond human priors, by post-training alone and with no added parameters, including a doom-loop regression in both runs that all 17 benchmarks are structurally unable to observe.


{
    \small
    \bibliographystyle{ieeenat_fullname}
    \bibliography{main,nanoVLM,nanoVLM_v2,sectionD_citation_additions}

\begin{thebibliography}{62}
\providecommand{\natexlab}[1]{#1}
\providecommand{\url}[1]{\texttt{#1}}
\expandafter\ifx\csname urlstyle\endcsname\relax
  \providecommand{\doi}[1]{doi: #1}\else
  \providecommand{\doi}{doi: \begingroup \urlstyle{rm}\Url}\fi

\bibitem[Ben~Allal et~al.(2025)Ben~Allal, Lozhkov, Bakouch, et~al.]{allal2025smollm2}
Loubna Ben~Allal, Anton Lozhkov, Elie Bakouch, et~al.
\newblock {SmolLM2}: When smol goes big -- data-centric training of a small language model.
\newblock In \emph{Second Conference on Language Modeling}, 2025.

\bibitem[Chee et~al.(2026)Chee, Lee, and Hsu]{chee2026dream}
Evelyn Chee, Mong~Li Lee, and Wynne Hsu.
\newblock {DREAM}: Dynamic prompts and guidedmix for efficient continual adaptation of visual-language models.
\newblock In \emph{WACV}, pages 5853--5863, 2026.

\bibitem[Chen et~al.(2024{\natexlab{a}})Chen, Li, Dong, Zhang, Zang, et~al.]{chen2024mmstar}
Lin Chen, Jinsong Li, Xiaoyi Dong, Pan Zhang, Yuhang Zang, et~al.
\newblock Are we on the right way for evaluating large vision-language models?
\newblock In \emph{NeurIPS}, pages 27056--27087, 2024{\natexlab{a}}.

\bibitem[Chen et~al.(2024{\natexlab{b}})Chen, Wu, Wang, Su, Chen, Xing, Zhong, Zhang, Zhu, Lu, et~al.]{chen2024internvl}
Zhe Chen, Jiannan Wu, Wenhai Wang, Weijie Su, Guo Chen, Sen Xing, Muyan Zhong, Qinglong Zhang, Xizhou Zhu, Lewei Lu, et~al.
\newblock Internvl: Scaling up vision foundation models and aligning for generic visual-linguistic tasks.
\newblock In \emph{CVPR}, pages 24185--24198, 2024{\natexlab{b}}.

\bibitem[Chu et~al.(2023)Chu, Qiao, Lin, Xu, Yang, Hu, Wei, Zhang, Zhang, Wei, and Shen]{chu2023mobilevlm}
Xiangxiang Chu, Limeng Qiao, Xinyang Lin, Shuang Xu, Yang Yang, Yiming Hu, Fei Wei, Xinyu Zhang, Bo Zhang, Xiaolin Wei, and Chunhua Shen.
\newblock {MobileVLM}: A fast, strong and open vision language assistant for mobile devices.
\newblock \emph{arXiv preprint arXiv:2312.16886}, 2023.

\bibitem[Das et~al.(2025)Das, Talon, Mancini, Wang, and Ricci]{das2025onevlm}
Deepayan Das, Davide Talon, Massimiliano Mancini, Yiming Wang, and Elisa Ricci.
\newblock One {VLM} to keep it learning: Generation and balancing for data-free continual visual question answering.
\newblock In \emph{WACV}, pages 5635--5645, 2025.

\bibitem[Dayal et~al.(2026)Dayal, Divya, Tiwari, Cenkeramaddi, Mohan, and Kumar]{dayal2026duaa}
Aveen Dayal, Peketi Divya, Nidhi Tiwari, Linga~Reddy Cenkeramaddi, C~Krishna Mohan, and Abhinav Kumar.
\newblock Bridging the domain gap in small multimodal models: A dual-level alignment perspective.
\newblock In \emph{WACV}, pages 8262--8271, 2026.

\bibitem[Dhouib et~al.(2025)Dhouib, Buscaldi, Vanier, and Shabou]{dhouib2025pact}
Mohamed Dhouib, Davide Buscaldi, Sonia Vanier, and Aymen Shabou.
\newblock Pact: Pruning and clustering-based token reduction for faster visual language models.
\newblock In \emph{CVPR}, pages 14582--14592, 2025.

\bibitem[Ding et~al.(2025)Ding, Wu, Zhao, Zang, Duan, Dong, Zhang, Cao, Lin, and Wang]{mmif23k_iccv25}
Shengyuan Ding, Shenxi Wu, Xiangyu Zhao, Yuhang Zang, Haodong Duan, Xiaoyi Dong, Pan Zhang, Yuhang Cao, Dahua Lin, and Jiaqi Wang.
\newblock Mm-ifengine: Towards multimodal instruction following.
\newblock In \emph{ICCV}, pages 1099--1109, 2025.

\bibitem[Duan et~al.(2024)Duan, Yang, Qiao, Fang, Chen, et~al.]{duan2024vlmevalkit}
Haodong Duan, Junming Yang, Yuxuan Qiao, Xinyu Fang, Lin Chen, et~al.
\newblock {VLMEvalKit}: An open-source toolkit for evaluating large multi-modality models.
\newblock In \emph{ACMMM}, pages 11198--11201, 2024.

\bibitem[Fu et~al.(2025)Fu, Chen, Shen, Qin, Zhang, Lin, Yang, Zheng, Li, Sun, et~al.]{fu2026mme}
Chaoyou Fu, Peixian Chen, Yunhang Shen, Yulei Qin, Mengdan Zhang, Xu Lin, Jinrui Yang, Xiawu Zheng, Ke Li, Xing Sun, et~al.
\newblock Mme: A comprehensive evaluation benchmark for multimodal large language models.
\newblock In \emph{NeurIPS}, 2025.

\bibitem[Gao et~al.(2025)Gao, Pi, Zhang, Ye, Zhong, Wang, Hong, Han, Xu, Li, and Kong]{gao2025gllava}
Jiahui Gao, Renjie Pi, Jipeng Zhang, Jiacheng Ye, Wanjun Zhong, Yufei Wang, Lanqing Hong, Jianhua Han, Hang Xu, Zhenguo Li, and Lingpeng Kong.
\newblock G-llava: Solving geometric problem with multi-modal large language model.
\newblock In \emph{International Conference on Learning Representations}, 2025.

\bibitem[Holtzman et~al.(2020)Holtzman, Buys, Du, Forbes, and Choi]{doomloop1}
Ari Holtzman, Jan Buys, Li Du, Maxwell Forbes, and Yejin Choi.
\newblock The curious case of neural text degeneration.
\newblock In \emph{ICLR}, 2020.

\bibitem[Ilharco et~al.(2023)Ilharco, Ribeiro, Wortsman, Gururangan, Schmidt, Hajishirzi, and Farhadi]{ilharco2023taskarithmetic}
Gabriel Ilharco, Marco~Tulio Ribeiro, Mitchell Wortsman, Suchin Gururangan, Ludwig Schmidt, Hannaneh Hajishirzi, and Ali Farhadi.
\newblock Editing models with task arithmetic.
\newblock In \emph{ICLR}, 2023.

\bibitem[Ji et~al.(2025)Ji, Chen, Pan, Zhu, Li, Hong, Chen, Zhou, Wang, Dai, Chan, Han, Guo, and Yang]{safety_data}
Jiaming Ji, Xinyu Chen, Rui Pan, Han Zhu, Jiahao Li, Donghai Hong, Boyuan Chen, Jiayi Zhou, Kaile Wang, Juntao Dai, Chi-Min Chan, Sirui Han, Yike Guo, and Yaodong Yang.
\newblock Safe rlhf-v: Safe reinforcement learning from multi-modal human feedback.
\newblock In \emph{NeurIPS}, 2025.

\bibitem[Jiang et~al.(2026)Jiang, Zhang, Zhao, Ma, Hou, Wu, Cai, Hu, Cheng, Chan, Xue, and Guo]{jiang2026fisa}
Chunyang Jiang, Pingping Zhang, Yuzhi Zhao, Wenao Ma, Zhijian Hou, Mengyang Wu, Yiyang Cai, Senkang Hu, Sitong Cheng, Chi-Min Chan, Wei Xue, and Yike Guo.
\newblock Failure-informed image self-augmentation for multimodal large language model self-improvement.
\newblock \emph{arXiv preprint arXiv:2608.03733}, 2026.

\bibitem[Kantharaj et~al.(2022)Kantharaj, Leong, Lin, Masry, Thakkar, Hoque, and Joty]{kantharaj2022charttotext}
Shankar Kantharaj, Rixie~Tiffany Leong, Xiang Lin, Ahmed Masry, Megh Thakkar, Enamul Hoque, and Shafiq Joty.
\newblock Chart-to-text: A large-scale benchmark for chart summarization.
\newblock In \emph{Proceedings of the 60th Annual Meeting of the Association for Computational Linguistics (Volume 1: Long Papers)}, pages 4005--4023, 2022.

\bibitem[Kembhavi et~al.(2016)Kembhavi, Salvato, Kolve, Seo, Hajishirzi, and Farhadi]{kembhavi2016ai2d}
Aniruddha Kembhavi, Mike Salvato, Eric Kolve, Minjoon Seo, Hannaneh Hajishirzi, and Ali Farhadi.
\newblock A diagram is worth a dozen images.
\newblock In \emph{ECCV}, pages 235--251, 2016.

\bibitem[Lab(2026)]{OpenRT2026}
Shanghai~AI Lab.
\newblock Openrt: An open-source red teaming framework for multimodal llms.
\newblock \emph{arXiv preprint arXiv:2601.01592}, 2026.

\bibitem[Lauren\c{c}on et~al.(2024)Lauren\c{c}on, Marafioti, Sanh, and Tronchon]{laurencon2024building}
Hugo Lauren\c{c}on, Andr\'{e}s Marafioti, Victor Sanh, and L\'{e}o Tronchon.
\newblock Building and better understanding vision-language models: Insights and future directions, 2024.

\bibitem[Li et~al.(2024{\natexlab{a}})Li, Ge, Ge, Wang, Wang, Zhang, and Shan]{li2023seedbench}
Bohao Li, Yuying Ge, Yixiao Ge, Guangzhi Wang, Rui Wang, Ruimao Zhang, and Ying Shan.
\newblock Seed-bench: Benchmarking multimodal large language models.
\newblock In \emph{CVPR}, pages 13299--13308, 2024{\natexlab{a}}.

\bibitem[Li et~al.(2024{\natexlab{b}})Li, Li, Yin, Ahmed, Liu, and Liu]{li-etal-2024-red}
Mukai Li, Lei Li, Yuwei Yin, Masood Ahmed, Zhenguang Liu, and Qi Liu.
\newblock Red teaming visual language models.
\newblock In \emph{Findings of the Association for Computational Linguistics: ACL 2024}, pages 3326--3342, Bangkok, Thailand, 2024{\natexlab{b}}. Association for Computational Linguistics.

\bibitem[Li et~al.(2023)Li, Du, Zhou, Wang, Zhao, and Wen]{li2023pope}
Yifan Li, Yifan Du, Kun Zhou, Jinpeng Wang, Wayne~Xin Zhao, and Ji-Rong Wen.
\newblock Evaluating object hallucination in large vision-language models.
\newblock In \emph{EMNLP}, pages 292--305, 2023.

\bibitem[{Liquid AI}(2025)]{liquid2025lfm2}
{Liquid AI}.
\newblock {LFM2} technical report.
\newblock \emph{arXiv preprint arXiv:2511.23404}, 2025.

\bibitem[{Liquid AI}(2026)]{liquid2026lfm25vl}
{Liquid AI}.
\newblock {LFM2.5-VL-450M}.
\newblock Hugging Face model card, 2026.

\bibitem[Liu et~al.(2023)Liu, Li, Wu, and Lee]{liu2023llava}
Haotian Liu, Chunyuan Li, Qingyang Wu, and Yong~Jae Lee.
\newblock Visual instruction tuning.
\newblock In \emph{NeurIPS}, pages 34892--34916, 2023.

\bibitem[Liu et~al.(2024{\natexlab{a}})Liu, Duan, Zhang, Li, Zhang, et~al.]{liu2024mmbench}
Yuan Liu, Haodong Duan, Yuanhan Zhang, Bo Li, Songyang Zhang, et~al.
\newblock {MMBench}: Is your multi-modal model an all-around player?
\newblock In \emph{ECCV}, pages 216--233, 2024{\natexlab{a}}.

\bibitem[Liu et~al.(2024{\natexlab{b}})Liu, Li, Huang, Yang, Yu, et~al.]{liu2024ocrbench}
Yuliang Liu, Zhang Li, Mingxin Huang, Biao Yang, Wenwen Yu, et~al.
\newblock {OCRBench}: On the hidden mystery of {OCR} in large multimodal models.
\newblock \emph{Science China Information Sciences}, 67\penalty0 (12), 2024{\natexlab{b}}.

\bibitem[Lu et~al.(2022)Lu, Mishra, Xia, Qiu, Chang, et~al.]{lu2022scienceqa}
Pan Lu, Swaroop Mishra, Tony Xia, Liang Qiu, Kai-Wei Chang, et~al.
\newblock Learn to explain: Multimodal reasoning via thought chains for science question answering.
\newblock In \emph{NeurIPS}, pages 2507--2521, 2022.

\bibitem[Lu et~al.(2024)Lu, Bansal, Xia, Liu, Li, et~al.]{lu2024mathvista}
Pan Lu, Hritik Bansal, Tony Xia, Jiacheng Liu, Chunyuan Li, et~al.
\newblock {MathVista}: Evaluating mathematical reasoning of foundation models in visual contexts.
\newblock In \emph{ICLR}, pages 23439--23554, 2024.

\bibitem[Marafioti et~al.(2025)Marafioti, Zohar, Farr\'e, Noyan, Bakouch, Cuenca, Zakka, Ben~Allal, Lozhkov, Tazi, Srivastav, Lochner, Larcher, Morlon, Tunstall, von Werra, and Wolf]{marafioti2025smolvlm}
Andr\'es Marafioti, Orr Zohar, Miquel Farr\'e, Merve Noyan, Elie Bakouch, Pedro Cuenca, Cyril Zakka, Loubna Ben~Allal, Anton Lozhkov, Nouamane Tazi, Vaibhav Srivastav, Joshua Lochner, Hugo Larcher, Mathieu Morlon, Lewis Tunstall, Leandro von Werra, and Thomas Wolf.
\newblock {SmolVLM}: Redefining small and efficient multimodal models.
\newblock \emph{arXiv preprint arXiv:2504.05299}, 2025.

\bibitem[Masry et~al.(2022)Masry, Long, Tan, Joty, and Hoque]{masry2022chartqa}
Ahmed Masry, Do~Xuan Long, Jia~Qing Tan, Shafiq Joty, and Enamul Hoque.
\newblock {ChartQA}: A benchmark for question answering about charts with visual and logical reasoning.
\newblock In \emph{Findings of ACL}, pages 2263--2279, 2022.

\bibitem[Mathew et~al.(2021)Mathew, Karatzas, and Jawahar]{mathew2021docvqa}
Minesh Mathew, Dimosthenis Karatzas, and C.~V. Jawahar.
\newblock {DocVQA}: A dataset for {VQA} on document images.
\newblock In \emph{WACV}, pages 2200--2209, 2021.

\bibitem[Mathew et~al.(2022)Mathew, Bagal, Tito, Karatzas, Valveny, and Jawahar]{mathew2022infographicvqa}
Minesh Mathew, Viraj Bagal, Rub\`en Tito, Dimosthenis Karatzas, Ernest Valveny, and C.~V. Jawahar.
\newblock {InfographicVQA}.
\newblock In \emph{WACV}, pages 1697--1706, 2022.

\bibitem[{NVIDIA}(2025)]{nvidia2025nemotron}
{NVIDIA}.
\newblock {Nemotron} image training dataset v3.
\newblock \url{https://huggingface.co/datasets/nvidia/Nemotron-Image-Training-v3}, 2025.

\bibitem[{Qwen Team}(2026{\natexlab{a}})]{qwen2026qwen35}
{Qwen Team}.
\newblock {Qwen3.5-0.8B}.
\newblock Hugging Face model card, 2026{\natexlab{a}}.

\bibitem[{Qwen Team}(2026{\natexlab{b}})]{qwen2026qwen36}
{Qwen Team}.
\newblock {Qwen3.6-27B}.
\newblock Hugging Face model card, 2026{\natexlab{b}}.

\bibitem[Rolnick et~al.(2019)Rolnick, Ahuja, Schwarz, Lillicrap, and Wayne]{rolnick2019replay}
David Rolnick, Arun Ahuja, Jonathan Schwarz, Timothy~P. Lillicrap, and Greg Wayne.
\newblock Experience replay for continual learning.
\newblock In \emph{NeurIPS}, 2019.

\bibitem[Sabbaghi et~al.(2026)Sabbaghi, Pappas, Javanmard, and Hassani]{sabbaghi2026infosft}
Mahdi Sabbaghi, George Pappas, Adel Javanmard, and Hamed Hassani.
\newblock {InfoSFT}: Learn more and forget less with information-aware token weighting.
\newblock \emph{arXiv preprint arXiv:2605.14967}, 2026.

\bibitem[Sarch et~al.(2026)Sarch, Cai, Wang, Wu, Chen, and Liu]{vero600k}
Gabriel Sarch, Linrong Cai, Qunzhong Wang, Haoyang Wu, Danqi Chen, and Zhuang Liu.
\newblock Vero: An open rl recipe for general visual reasoning.
\newblock \emph{arXiv preprint arXiv:2604.04917}, 2026.

\bibitem[Shi et~al.(2024)Shi, Hu, Bin, Liu, Yang, Ng, Bing, and Lee]{shi2024mathllava}
Wenhao Shi, Zhiqiang Hu, Yi Bin, Junhua Liu, Yang Yang, See-Kiong Ng, Lidong Bing, and Roy Ka-Wei Lee.
\newblock Math-llava: Bootstrapping mathematical reasoning for multimodal large language models.
\newblock \emph{arXiv preprint arXiv:2406.17294}, 2024.

\bibitem[Singh et~al.(2019)Singh, Natarajan, Shah, Jiang, Chen, et~al.]{singh2019textvqa}
Amanpreet Singh, Vivek Natarajan, Meet Shah, Yu Jiang, Xinlei Chen, et~al.
\newblock Towards {VQA} models that can read.
\newblock In \emph{CVPR}, pages 8317--8326, 2019.

\bibitem[Team et~al.(2025)Team, Kamath, Ferret, Pathak, Vieillard, Merhej, Perrin, Matejovicova, Ram{\'e}, Rivi{\`e}re, et~al.]{team2025gemma}
Gemma Team, Aishwarya Kamath, Johan Ferret, Shreya Pathak, Nino Vieillard, Ramona Merhej, Sarah Perrin, Tatiana Matejovicova, Alexandre Ram{\'e}, Morgane Rivi{\`e}re, et~al.
\newblock Gemma 3 technical report.
\newblock \emph{arXiv preprint arXiv:2503.19786}, 2025.

\bibitem[Tschannen et~al.(2025)Tschannen, Gritsenko, Wang, Naeem, Alabdulmohsin, et~al.]{tschannen2025siglip2}
Michael Tschannen, Alexey Gritsenko, Xiao Wang, Muhammad~Ferjad Naeem, Ibrahim Alabdulmohsin, et~al.
\newblock {SigLIP 2}: Multilingual vision-language encoders with improved semantic understanding, localization, and dense features.
\newblock \emph{arXiv preprint arXiv:2502.14786}, 2025.

\bibitem[Vasu et~al.(2025)Vasu, Faghri, Li, Koc, True, Antony, Santhanam, Gabriel, Grasch, Tuzel, and Pouransari]{vasu2024fastvlm}
Pavan Kumar~Anasosalu Vasu, Fartash Faghri, Chun-Liang Li, Cem Koc, Nate True, Albert Antony, Gokula Santhanam, James Gabriel, Peter Grasch, Oncel Tuzel, and Hadi Pouransari.
\newblock Fastvlm: Efficient vision encoding for vision language models.
\newblock In \emph{CVPR}, pages 19769--19780, 2025.

\bibitem[Wang et~al.(2024)Wang, Chen, Wang, Cao, Liu, et~al.]{wang2024mpo}
Weiyun Wang, Zhe Chen, Wenhai Wang, Yue Cao, Yangzhou Liu, et~al.
\newblock Enhancing the reasoning ability of multimodal large language models via mixed preference optimization.
\newblock \emph{arXiv preprint arXiv:2411.10442}, 2024.

\bibitem[Wei et~al.(2026)Wei, Cheng, Jin, Yang, Shen, Hou, Du, Yuan, Cao, and Tao]{wei2025optmerge}
Yongxian Wei, Runxi Cheng, Weike Jin, Enneng Yang, Li Shen, Lu Hou, Sinan Du, Chun Yuan, Xiaochun Cao, and Dacheng Tao.
\newblock {OptMerge}: Unifying multimodal {LLM} capabilities and modalities via model merging.
\newblock \emph{ICLR}, pages 61571--61595, 2026.

\bibitem[Welleck et~al.(2020)Welleck, Kulikov, Roller, Dinan, Cho, and Weston]{doomloop2}
Sean Welleck, Ilia Kulikov, Stephen Roller, Emily Dinan, Kyunghyun Cho, and Jason Weston.
\newblock Neural text generation with unlikelihood training.
\newblock In \emph{ICLR}, 2020.

\bibitem[Wiedmann et~al.(2025{\natexlab{a}})Wiedmann, Gosthipaty, and Marafioti]{huggingface2025nanovlm}
Luis Wiedmann, Aritra~Roy Gosthipaty, and Andrés Marafioti.
\newblock nanovlm.
\newblock \url{https://github.com/huggingface/nanoVLM}, 2025{\natexlab{a}}.

\bibitem[Wiedmann et~al.(2025{\natexlab{b}})Wiedmann, Zohar, Mahla, Wang, Li, Frere, von Werra, Roy~Gosthipaty, and Marafioti]{finevision2025}
Luis Wiedmann, Orr Zohar, Amir Mahla, Xiaohan Wang, Rui Li, Thibaud Frere, Leandro von Werra, Aritra Roy~Gosthipaty, and Andr{\'e}s Marafioti.
\newblock {FineVision}: Open data is all you need.
\newblock \emph{arXiv preprint arXiv:2510.17269}, 2025{\natexlab{b}}.

\bibitem[Wortsman et~al.(2022)Wortsman, Ilharco, Gadre, Roelofs, Gontijo-Lopes, et~al.]{wortsman2022soups}
Mitchell Wortsman, Gabriel Ilharco, Samir~Yitzhak Gadre, Rebecca Roelofs, Raphael Gontijo-Lopes, et~al.
\newblock Model soups: Averaging weights of multiple fine-tuned models improves accuracy without increasing inference time.
\newblock In \emph{ICML}, pages 23965--23998, 2022.

\bibitem[{xAI}(2024)]{xai2024realworldqa}
{xAI}.
\newblock {RealWorldQA}.
\newblock \url{https://huggingface.co/datasets/xai-org/RealworldQA}, 2024.

\bibitem[Xiong et~al.(2025)Xiong, Wang, Guo, Ye, Fan, Gu, Huang, and Li]{xiong2025llavacritic}
Tianyi Xiong, Xiyao Wang, Dong Guo, Qinghao Ye, Haoqi Fan, Quanquan Gu, Heng Huang, and Chunyuan Li.
\newblock Llava-critic: Learning to evaluate multimodal models.
\newblock In \emph{CVPR}, pages 13618--13628, 2025.

\bibitem[Yadav et~al.(2023)Yadav, Tam, Choshen, Raffel, and Bansal]{yadav2023ties}
Prateek Yadav, Derek Tam, Leshem Choshen, Colin Raffel, and Mohit Bansal.
\newblock {TIES-Merging}: Resolving interference when merging models.
\newblock In \emph{NeurIPS}, pages 7093--7115, 2023.

\bibitem[Yang et~al.(2025)Yang, Patel, Deitke, Gupta, Weihs, Head, Yatskar, Callison-Burch, Krishna, Kembhavi, and Clark]{yang2025cosyn}
Yue Yang, Ajay Patel, Matt Deitke, Tanmay Gupta, Luca Weihs, Andrew Head, Mark Yatskar, Chris Callison-Burch, Ranjay Krishna, Aniruddha Kembhavi, and Christopher Clark.
\newblock Scaling text-rich image understanding via code-guided synthetic multimodal data generation.
\newblock In \emph{Proceedings of the 63rd Annual Meeting of the Association for Computational Linguistics (Volume 1: Long Papers)}, pages 17486--17505, 2025.

\bibitem[Yao et~al.(2025)Yao, Wang, and Huang]{yao2024errordriven}
Barry~Menglong Yao, Qifan Wang, and Lifu Huang.
\newblock Error-driven data-efficient large multimodal model tuning.
\newblock In \emph{ACL}, pages 20289--20306, 2025.

\bibitem[Ye et~al.(2025)Ye, Gan, Ge, Zhang, and Tang]{ye2025atpllava}
Xubing Ye, Yukang Gan, Yixiao Ge, Xiao-Ping Zhang, and Yansong Tang.
\newblock Atp-llava: Adaptive token pruning for large vision language models.
\newblock In \emph{CVPR}, pages 24972--24982, 2025.

\bibitem[Yu et~al.(2026)Yu, Wang, Wang, Huang, Ma, He, Cai, Chen, Huang, Zhao, et~al.]{yu2026minicpm}
Tianyu Yu, Zefan Wang, Chongyi Wang, Fuwei Huang, Wenshuo Ma, Zhihui He, Tianchi Cai, Weize Chen, Yuxiang Huang, Ranchi Zhao, et~al.
\newblock Minicpm-v 4.5: Cooking efficient mllms via architecture, data, and training recipe.
\newblock In \emph{CVPR}, pages 11704--11715, 2026.

\bibitem[Yu et~al.(2024)Yu, Yang, Li, Wang, Lin, et~al.]{yu2023mmvet}
Weihao Yu, Zhengyuan Yang, Linjie Li, Jianfeng Wang, Kevin Lin, et~al.
\newblock {MM-Vet}: Evaluating large multimodal models for integrated capabilities.
\newblock \emph{ICML}, pages 57730--57754, 2024.

\bibitem[Yue et~al.(2024)Yue, Ni, Zhang, Zheng, Liu, et~al.]{yue2024mmmu}
Xiang Yue, Yuansheng Ni, Kai Zhang, Tianyu Zheng, Ruoqi Liu, et~al.
\newblock {MMMU}: A massive multi-discipline multimodal understanding and reasoning benchmark for expert {AGI}.
\newblock In \emph{CVPR}, pages 9556--9567, 2024.

\bibitem[Zamini and Shukla(2026)]{zamini2026deltallava}
Mohamad Zamini and Diksha Shukla.
\newblock Delta-llava: Base-then-specialize alignment for token-efficient vision-language models.
\newblock In \emph{WACV}, pages 3648--3657, 2026.

\bibitem[Zhang et~al.(2024)Zhang, Hu, Xu, Yan, Xu, Jin, Zhang, and Huang]{zhang2024tinychart}
Liang Zhang, Anwen Hu, Haiyang Xu, Ming Yan, Yichen Xu, Qin Jin, Ji Zhang, and Fei Huang.
\newblock Tinychart: Efficient chart understanding with program-of-thoughts learning and visual token merging.
\newblock In \emph{Proceedings of the 2024 Conference on Empirical Methods in Natural Language Processing}, pages 1882--1898, 2024.

\end{thebibliography}
}
\clearpage
\appendix
\begin{center}
    {\LARGE\bfseries Appendix}
\end{center}

This appendix provides additional details and results.

\vspace{0.6em}


\begin{center}
    {\large\bfseries Table of Contents}
\end{center}

\begingroup
\small
\setlength{\parskip}{0pt}

\noindent
\textbf{A. Details of the Active Failure-Discovery Harness}
\dotfill \pageref{app:activefailure}\\
\hspace*{1.5em}A.1 Harness Configuration
\dotfill \pageref{app:harness_config}\\
\hspace*{1.5em}A.2 Loop Mechanics
\dotfill \pageref{app:loop_mechanics}\\
\hspace*{1.5em}A.3 Seed Catalogue
\dotfill \pageref{app:seed_catalogue}\\
\hspace*{1.5em}A.4 How the Loop Refined the Seeds
\dotfill \pageref{app:seed_refinement}\\
\hspace*{1.5em}A.5 Novel Weaknesses Discovered by the Teacher
\dotfill \pageref{app:novel_weaknesses}\\

\noindent
\textbf{B. Extended On-Device Evaluation}
\dotfill \pageref{app:on_device}\\
\hspace*{1.5em}B.1 Quantization and Accuracy Retention
\dotfill \pageref{app:quantization}\\
\hspace*{1.5em}B.2 Evaluation Protocol
\dotfill \pageref{app:on_device_protocol}\\
\hspace*{1.5em}B.3 Visual-Token Footprint
\dotfill \pageref{app:visual_tokens}\\
\hspace*{1.5em}B.4 Per-Backend TTFT Profiling
\dotfill \pageref{app:backend_results}\\
\hspace*{1.5em}B.5 Peak Memory
\dotfill \pageref{app:peak_memory}\\
\hspace*{1.5em}B.6 Autoregressive Decode Throughput
\dotfill \pageref{app:decode_throughput}\\
\hspace*{1.5em}B.7 End-to-End Latency
\dotfill \pageref{app:ttlt}\\[0.45em]

\noindent
\textbf{C. Additional Ablations}
\dotfill \pageref{app:additional_ablations}\\
\hspace*{1.5em}C.1 Robustness to the Choice of LLM Judge
\dotfill \pageref{app:judge_robustness}\\
\hspace*{1.5em}C.2 Impact of Post-Training under the Bounded Native-Resolution Policy
\dotfill \pageref{app:sft_ablation}\\[0.45em]

\noindent
\textbf{D. Implementation Details}
\dotfill \pageref{app:implementation_details}\\
\hspace*{1.5em}D.1 Evaluation Protocol
\dotfill \pageref{app:eval_protocol}\\
\hspace*{1.5em}D.2 Training Details
\dotfill \pageref{app:stage1_ID}\\[0.45em]

\noindent
\textbf{E. Qualitative Results and On-Device Deployment}
\dotfill \pageref{app:qualitative_appendix}\\
\hspace*{1.5em}E.1 Qualitative Comparisons 
\dotfill \pageref{app:qualitative_results}\\
\hspace*{1.5em}E.2 Failure Cases and Limitations 
\dotfill \pageref{app:qualitative_failures}\\
\hspace*{1.5em}E.3 Live On-Device Deployment
\dotfill \pageref{app:on_device_deployment}

\endgroup

\vspace{0.8em}

\section{Details of the active failure-discovery harness}
\label{app:activefailure}

\subsection{Harness configuration}
\label{app:harness_config}
The teacher is Qwen3.6-27B served in FP8, and the student $\theta_2$ is served
through the identical inference stack used for benchmark evaluation, so every probe
response is exactly what the model would produce under test. A single $8$-GPU node
hosts the pair, with the teacher replicated across seven GPUs under vLLM data
parallelism and the student on the eighth; $21$ images are processed concurrently per
round. Each round samples $21$ images and issues up to $4$ probes per image, and
Reflect runs at the end of every round. Planning uses temperature $1.0$ to keep probes
diverse, while judging and reflection use temperature $0.6$ with an $8192$-token
budget, which is what allows the teacher to reason before emitting strict JSON.
Persistent state lives in three artifacts: the weakness catalogue, an append-only log
of every probe with its verdict, and per-source counters that drive sampling,
alongside per-category counters for reporting.

\subsection{Loop mechanics}
\label{app:loop_mechanics}
\paragraph{Plan.} The teacher receives the image, the dataset's reference Q\&A, and a
summary of the current catalogue, and returns $k$ probes as strict JSON. Each probe
carries an expected answer, a capability category, a free-form persona naming who
would plausibly ask it, and a tag naming the known weakness it is meant to stress.
Feeding the catalogue back in at this point is what makes the loop adversarial rather
than exploratory: the planner deliberately re-attacks soft spots instead of asking
generic questions.

\paragraph{Answer.} The student answers each probe independently through the standard
pipeline, with no prompt engineering, retries, or decoding changes.

\paragraph{Judge.} The teacher grades all $k$ answers for an image in a single call
with the image attached, returning a verdict, a severity in $[1,5]$, a category, and a
short rationale per answer. The rubric names the pathologies we care about explicitly:
an answer that repeats a word, phrase, or line until it is cut off is a failure at
severity $5$ regardless of whether its opening was correct; character-level and
tokenization slips are labelled as generation idiosyncrasies; and empty, near-empty,
or pure-filler answers are labelled as refusals. Verdicts are \emph{pass},
\emph{partial}, or \emph{fail}. When the teacher cannot return parseable output even
after a retry with reasoning disabled, the answers are recorded as \emph{ungraded}
rather than guessed, and ungraded records are excluded from every rate we report and
from reflection, so infrastructure noise can never inflate a failure statistic.

\paragraph{Reflect.} At the end of the round the teacher receives the current
catalogue together with the $40$ most recent failures and returns the full catalogue
rewritten: entries may be merged, re-described, re-ranked in severity, or created.
Every failure is mapped onto a fixed taxonomy of $19$ labels ($18$ capability
categories plus \emph{other}), which is what keeps the catalogue aggregable across
hundreds of rounds instead of sprawling into near-duplicate clusters. Per-cluster
evidence counts are recomputed from the probe log rather than taken from the teacher,
so the statistics remain grounded even when its prose drifts. The rewritten catalogue
is exactly what the next round's planner consumes, closing the loop.

\subsection{Seed catalogue}
\label{app:seed_catalogue}

\paragraph{Why seed at all.} Grounding failure discovery in a predefined rubric is
standard practice: RTVLM~\cite{li-etal-2024-red} fixes four red-teaming aspects over twelve
task categories and attaches per-category scoring criteria to each. Our seed catalogue
plays that role, steering the loop toward behaviors we already suspected and making it
adversarial from round~$1$ rather than spending its first rounds rediscovering what we
knew. It also keeps the comparison honest, since the claim of going beyond human priors
means little unless the prior is the strongest one we could write down. The difference
is what happens next: a red-teaming rubric is fixed for the lifetime of the benchmark,
whereas ours is an initial condition the teacher may re-rank, rewrite, or
abandon, and over $300$ rounds it did all three (App.~\ref{app:seed_refinement}).

\paragraph{How the seeds were written.} The $13$ seeds of Table~\ref{tab:seeds} come
from three sources. The first is an error analysis of $\theta_2$ across the four
capability areas, grouped by question type, which localises a weakness to specific
slices instead of an aggregate score. We repeated that analysis on the base nanoVLM checkpoint and on two other $\sim$0.5B models, SmolVLM2-500M and LFM2.5-VL-450M, which
sorts each weak slice into a kind. Where every model is weak, the difficulty is
intrinsic to the scale. Where only ours is weak, the training mixture is at fault.
Where a competitor already succeeds, the capability is learnable at $460$M, which makes
it the most promising thing to seed. The second source is manual inspection of the
failing generations in those slices: reading outputs rather than scores is what named
the digit-OCR and doom-loop entries. The third is failure modes generic to sub-billion
VLMs that our own inspection corroborated.

Each seed is written in the schema the teacher emits during Reflect, so the planner
cannot distinguish an operator entry from one the loop authored later. 

\begin{table}[htbp]
    \centering
    \footnotesize
    \begin{tabularx}{\linewidth}{@{}
        >{\raggedright\arraybackslash\hsize=0.72\hsize}X
        >{\raggedright\arraybackslash\hsize=1.28\hsize}X @{}}
        \toprule
        \textbf{Seed weakness (category)} & \textbf{Seeded hypothesis} \\
        \midrule
        Doom-loop / degenerate repetition \textit{(repetition)} &
        On long generations the model repeats a word, phrase, or line until the token limit instead of stopping cleanly. \\
        \addlinespace
        Ignores output-format instructions \textit{(instruction following)} &
        Given a required output shape (JSON with fixed keys, one word, a bare number) it adds commentary or picks its own format. \\
        \addlinespace
        Fails exhaustive extraction \textit{(instruction following)} &
        Told to list \emph{all} text or items, it omits entries, stops early, or pads with commentary. \\
        \addlinespace
        Drops multi-part / conditional instructions \textit{(instruction following)} &
        With multiple steps or a condition (``if $X$ then $\dots$, else NONE'') it handles only the first part. \\
        \addlinespace
        Brittle to natural-language chart phrasing \textit{(chart reading)} &
        Reads chart values under templated phrasing but fails on paraphrased, conversational versions of the same lookup. \\
        \addlinespace
        Numeric reasoning over read values \textit{(math / numeric)} &
        Reads individual numbers correctly but cannot sum, difference, average, or count over them. \\
        \addlinespace
        Digit / dense-numeral OCR \textit{(OCR)} &
        Misreads prices, times, codes, and small or dense numerals; numeric scene text is far weaker than word reading. \\
        \addlinespace
        Multi-span synthesis \textit{(document understanding)} &
        Locates a single fact but cannot combine facts from different regions of a document into one answer. \\
        \addlinespace
        Structured-layout reading \textit{(document understanding)} &
        Weak on tables, forms, and handwriting relative to plain prose; loses row/column and field associations. \\
        \addlinespace
        Spatial reasoning \& OCR-spatial integration \textit{(spatial)} &
        Confuses left/right and above/below, mis-traces arrows, and attaches the wrong label to a region. \\
        \addlinespace
        Figure / diagram QA \textit{(diagram \& science)} &
        Struggles on questions over scientific figures, plots, and labelled diagrams. \\
        \addlinespace
        Fine-grained perception \textit{(fine-grained recognition)} &
        Cannot separate visually similar objects, icons, or subtle attribute differences. \\
        \addlinespace
        Geometry / multi-step math \textit{(math / numeric)} &
        Beyond single-step arithmetic: chained problems and geometric reasoning where intermediate results must be carried. \\
        \bottomrule
    \end{tabularx}
    \vspace{2mm}
    \caption{The $13$ seeded weaknesses that initialise the catalogue, derived from an error analysis of $\theta_2$, inspection of its failing generations, and failure modes generic to sub-billion VLMs. Seeds are hypotheses rather than ground truth: the loop re-estimates their severity, rewrites their descriptions, and replaces their probe templates from observed evidence.}
    \label{tab:seeds}
\end{table}
\subsection{How the loop refined the seeds}
\label{app:seed_refinement}
The seeds did not survive contact with evidence unchanged. The teacher rewrote all
$13$ hypotheses and replaced all $13$ probe template sets, and it re-ranked severity
rather than inheriting our estimates: two seeds were escalated to the maximum severity
(digit/dense-numeral OCR, and figure/diagram QA) and none were downgraded. One seed
the evidence did not support, geometry and multi-step math, received no targeted
probes and was effectively abandoned by the planner. Table~\ref{tab:refine} shows what
refinement buys: the operator's prior is a broad claim, while the refined entry
enumerates the specific confusions actually observed and collapses the probe templates
into short elicitations that had already proven capable of triggering the failure.
This matters downstream, because it is the refined description, not our original
guess, that names a concrete data-synthesis recipe. Grouped by capability area, the
confirmed and discovered clusters define the $K\!=\!6$ weakness-targeted mixtures of
Stage~3 (Sec.~3.2), with output formatting, exhaustive extraction, numeric reasoning
over read values, spatial--OCR integration, digit OCR, and doom-loop termination
carrying the largest accumulated evidence.

\begin{table}[htbp]
    \centering
    \footnotesize
    \begin{tabularx}{\linewidth}{@{} l >{\raggedright\arraybackslash}X @{}}
        \toprule
        \multicolumn{2}{@{}l}{\textbf{Seed:} Digit / dense-numeral OCR \quad severity $4 \rightarrow 5$} \\
        \midrule
        Seeded hypothesis & Misreads prices, times, codes, and small or dense numerals in scene text and documents (numeric scene text much weaker than word reading). \\
        \addlinespace
        Refined hypothesis & Misreads prices, times, codes, and dense numerals; high error rate on alphanumeric IDs; confuses similar digits; misses small text in footers and badges. New evidence: misreads numeric labels in geometry diagrams, omits the ``\%'' symbol, hallucinates license-plate digits, fails on axis tick marks, reads ``S\$6'' as ``56'' and ``35\%'' as ``39'', misreads scale-bar numbers and zip codes in footers. \\
        \addlinespace
        Seeded probes & ``Read the exact number/price/time shown here.''; ``Transcribe the full code/serial/plate exactly.'' \\
        \addlinespace
        Refined probes & ``Read the price on this tag.''; ``Transcribe the license plate.''; ``What time is on the clock?'' \\
        \bottomrule
    \end{tabularx}
    \vspace{2mm}
    \caption{Representative refinement of a seeded weakness after $300$ rounds. The teacher replaces a broad operator prior with an enumeration of observed mechanisms and distils the probe templates into short elicitations that reliably trigger the failure.}
    \label{tab:refine}
\end{table}

\subsection{Novel weaknesses discovered by the teacher}
\label{app:novel_weaknesses}
The study ran $300$ rounds against $\theta_2$, issuing $18{,}148$ probes drawn from
all $52$ FineVision\cite{finevision2025} sources, of which $17{,}811$ were graded. Of the graded probes,
$78.3\%$ were scored fail or partial ($12{,}430$ fail, $1{,}511$ partial, $3{,}870$
pass); the remaining $337$ were marked ungraded and are excluded from this and every
other rate we report. A uniform sampler over the same pool would look nothing like
this: the failure density is a direct consequence of conditioning the planner on the
catalogue and tilting sampling toward sources the student is currently failing, which
is what allows a fixed probe budget to buy diagnosis rather than confirmation of what
the model already does well.

Discovery is sharply front-loaded. The first novel cluster appears in round~$1$, half
of the $24$ are catalogued by round~$41$, and all $24$ by round~$51$; over rounds
$51$--$229$, the last round in which the catalogue changed, a further $10{,}872$
graded probes produced no new cluster, only additional evidence, re-ranked severities,
and sharpened hypotheses. In this run the catalogue therefore saturates within a few
dozen rounds, which bounds the compute required before a diagnosis is stable enough to
commit to a data mixture. Table~\ref{tab:novel} lists all $24$ teacher-authored
clusters with the round in which each entered the catalogue.

\begin{table}[htbp]
    \centering
    \footnotesize
    \begin{tabularx}{\linewidth}{@{} c c >{\raggedright\arraybackslash}X @{}}
        \toprule
        \textbf{Rd.} & \textbf{Sev.} & \textbf{Discovered failure mode (category)} \\
        \midrule
        1  & 5 & Hallucination of visual content and text \textit{(hallucination)}$^{\dagger}$ \\
        1  & 5 & Refusal or empty output on complex tasks \textit{(refusal)}$^{\dagger}$ \\
        2  & 5 & Attribute recognition errors \textit{(attributes)}$^{\dagger}$ \\
        3  & 5 & Chinese / complex-script OCR failure \textit{(OCR)} \\
        3  & 4 & Hallucinated errors when asked to spot textual mistakes \textit{(hallucination)}$^{\dagger}$ \\
        5  & 5 & Counting failures in cluttered scenes \textit{(counting)}$^{\dagger}$ \\
        6  & 5 & Grounding failure on spatial queries \textit{(grounding)}$^{\dagger}$ \\
        16 & 5 & World-knowledge hallucination in visual QA \textit{(world knowledge)}$^{\dagger}$ \\
        18 & 5 & Defaults to generic transcription on specific queries \textit{(instruction following)} \\
        19 & 5 & Cross-table / cross-region confusion in documents \textit{(document)} \\
        41 & 5 & Stem-and-leaf plot misinterpretation \textit{(chart reading)} \\
        41 & 5 & Gibberish output on simple arithmetic \textit{(generation idiosyncrasy)}$^{\dagger}$ \\
        42 & 5 & Chart axis-label OCR and extraction failure \textit{(OCR)} \\
        42 & 5 & Chart value reading under conditional logic \textit{(chart reading)} \\
        45 & 4 & Reading-order failure in multi-column text \textit{(OCR)} \\
        48 & 5 & Stacked-chart segment misidentification \textit{(chart reading)} \\
        48 & 5 & Cross-category data association error \textit{(chart reading)} \\
        48 & 5 & Table row/column misalignment during OCR \textit{(OCR)} \\
        48 & 5 & Conditional output-format failure \textit{(instruction following)} \\
        48 & 5 & Sub-text and multi-line text omission \textit{(OCR)} \\
        51 & 5 & Table category/role misclassification \textit{(document)} \\
        51 & 5 & Header versus data-row confusion in tables \textit{(document)} \\
        51 & 4 & Invalid JSON syntax generation \textit{(instruction following)} \\
        51 & 4 & Thought-process leakage into the answer \textit{(instruction following)} \\
        \bottomrule
    \end{tabularx}
    \vspace{2mm}
    \caption{All $24$ weaknesses authored by the teacher over $300$ rounds, with the round each first entered the catalogue and its final severity; $20$ of the $24$ carry the maximum severity, and together with the $13$ seeds they form a final catalogue of $37$ clusters. $^{\dagger}$ marks the seven categories no seed covered. Four of these seven (hallucination, refusal, world knowledge, generation idiosyncrasy) are absent from the seed set in any form; the other three (counting, attributes, grounding) name behaviors mentioned inside a seed hypothesis but never promoted to a category of their own. Entries without a dagger are novel mechanisms discovered within a seeded category, such as the resolution of a single ``digit OCR'' prior into axis-label extraction, row/column misalignment, and multi-column reading order.}
    \label{tab:novel}
\end{table}
\section{Extended On-Device Evaluation}
\label{app:on_device}

This section provides extended analyses for the on-device evaluation (\ref{sec:on_device}). We first characterize the accuracy-storage trade-off under weight quantization, followed by the complete experimental protocol, an analysis of the visual-token footprint and additional efficiency measurements: per-backend latency, peak memory, autoregressive decode throughput, and end-to-end latency.

\subsection{Quantization and Accuracy Retention}
\label{app:quantization}
We evaluate the robustness of GGUF checkpoints to weight quantization. Across all configurations, the vision encoder and multimodal projector are fixed to Q8\_0 (104\,MB), while only the language-model weights are compressed. Quantization labels therefore refer exclusively to the language model, whereas the reported total VLM size includes both components.

We evaluate each quantized checkpoint using the identical benchmark suite and scoring protocol from the main evaluation. To isolate quantization effects from intrinsic architectural differences, we report the change in normalized aggregate score ($\Delta$ Norm.) relative to the uncompressed FP32 GGUF reference of the corresponding checkpoint.

\begin{table}[ht]
    \centering
    \scriptsize
    \setlength{\tabcolsep}{6.0pt}
    \begin{tabular}{@{}lcccc@{}}
        \toprule
        LM format
        & Imatrix
        & VLM size (MB)
        & \textit{\Nano} $\Delta$ Norm.
        & \textit{\Flash} $\Delta$ Norm. \\
        \midrule
        FP32 & N/A & 1{,}666 & $0.00$ & $0.00$ \\
        BF16 & N/A & 886 & $-0.09$ & $-0.02$ \\
        Q8\_0 & No & 520 & $0.00$ & $-0.06$ \\
        Q5\_K\_M & No & 414 & $-0.12$ & $-0.52$ \\
        Q5\_K\_M & Yes & 414 & $-0.59$ & $-0.07$ \\
        Q4\_K\_M & Yes & 393 & $-0.24$ & $-0.53$ \\
        IQ4\_XS & Yes & 346 & $-0.92$ & $-0.91$ \\
        IQ4\_NL & Yes & 348 & $-0.96$ & $-0.97$ \\
        Q4\_0 & No & 348 & $-2.85$ & $-2.52$ \\
        IQ3\_M & Yes & 344 & $-1.36$ & $-1.57$ \\
        IQ3\_XXS & Yes & 334 & $-1.70$ & $-1.71$ \\
        \bottomrule
    \end{tabular}
    \caption{
        Accuracy retention across GGUF quantization formats.
    }
    \label{tab:quantization_extended}
\end{table}

Table~\ref{tab:quantization_extended} shows that both checkpoints retain their aggregate accuracy under calibrated low-bit quantization. The Q4\_K\_M format reduces the complete VLM footprint from 1{,}666 to 393\,MB (a $76.4\%$ reduction) while decreasing the normalized score by only $0.24$ points for \Nano and $0.53$ points for \Flash. The more compact IQ4 variants reduce the memory further to 346--348\,MB while remaining within one normalized point of their FP32 references.

Accuracy retention depends heavily on the quantization scheme rather than the raw storage footprint. For instance, Q4\_0 occupies approximately the same space as the IQ4 variants, yet incurs steeper penalties of $2.85$ points for \Nano and $2.52$ points for \Flash. Similarly, calibrated 3-bit variants preserve more accuracy than Q4\_0 despite smaller memory. The best calibrated operating point differs by checkpoint: Q4\_K\_M for VPsy and Q5\_K\_M for VPsy-Flash, each within 0.3 normalized points of its FP32 reference. We explicitly utilize Q4\_0 in our main cross-model experiments (Sec.~\ref{sec:on_device}) to ensure a standardized, hardware-agnostic comparison across all evaluated third-party architectures.

Finally, \Nano and \Flash exhibit nearly identical degradation trajectories across all formats. We conclude that training \Flash under a drastically reduced visual-token budget does not make its language model more sensitive to subsequent weight quantization. \Flash preserves its inference speedup without compromising quantization robustness.

\subsection{Evaluation Protocol}
\label{app:on_device_protocol}
Expanding on the experimental setup outlined in ~\cref{sec:on_device}, all models are evaluated using a modified fork of \texttt{llama.cpp} that introduces support for the architecture underlying \Fam and implements batched multimodal-projector execution. Following the quantization strategy detailed in~\cref{app:quantization}, all cross-model configurations utilize Q4\_0 language-model weights and Q8\_0 vision components.

Table~\ref{tab:device_setup} details the hardware specifications (SoC and memory capacity) and GPU compute backends for the evaluated smartphones. 

\begin{table}[ht]
    \centering
    \scriptsize
    \setlength{\tabcolsep}{4.0pt}
    \begin{tabular}{@{}lllr@{}}
        \toprule
        Device & SoC & GPU backend & RAM \\
        \midrule
        Google Pixel 9 & Tensor G4 & Vulkan & 12\,GB \\
        Samsung Galaxy S23 & Snapdragon 8 Gen 2 & OpenCL & 8\,GB \\
        Samsung Galaxy S25 Ultra & Snapdragon 8 Elite & OpenCL & 12\,GB \\
        Apple iPhone 15 & A16 Bionic & Metal & 6\,GB \\
        \bottomrule
    \end{tabular}
    \caption{Hardware specifications and compute backends for on-device evaluation.}
    \label{tab:device_setup}
\end{table}

While the main text reports the optimal hardware lane, complete CPU and GPU measurements for every model-device pair are provided in Sec.~\ref{app:backend_results}. To ensure rigorous reproducibility, the input image, textual prompt, and system configuration remain strictly fixed across all runs. Reported latencies represent warm medians over three timed runs following one harness-level warmup, strictly excluding model loading and runtime initialization. For controlled latency and throughput benchmarking, autoregressive generation is forced to exactly 32 output tokens by disabling early end-of-sequence termination. Time to first token is extracted directly from the runtime instrumentation as the sum of \texttt{encode\_ms} and \texttt{prefill\_ms}, accurately isolating the exact overhead of image encoding, multimodal-projector execution, and language-model prefill. End-to-end latency experiments under natural termination conditions are analyzed separately in Sec.~\ref{app:ttlt}. Finally, note that in our extended tabular results (\textit{e.g.}, \cref{tab:visual_tokens,tab:ttft_backend_full,tab:peak_memory_full,tab:decode_throughput_full}), we include Qwen3.5-0.8B as an additional baseline for completeness. We intentionally omitted this model from the main text comparisons for two reasons. First, its parameter count (${\sim}0.8$B) falls outside the strict efficiency-oriented target class (${\sim}0.5$B) evaluated in our work. Second, as our profiling reveals, its heavy visual-token footprint and architectural overhead render it highly inefficient for real-time edge deployment, making it unsuitable for a direct interactivity comparison with edge-native VLMs.

\subsection{Visual-Token Footprint}
\label{app:visual_tokens}

Table~\ref{tab:visual_tokens} reports the exact number of projected visual embeddings inserted into the language-model context, excluding textual delimiters, padding, and special tokens.

\begin{table}[ht]
    \centering
    \scriptsize
    \setlength{\tabcolsep}{4.0pt}
    \begin{tabular}{@{}lccc@{}}
        \toprule
        Model
        & $512^2$
        & $512{\times}1024$
        & $2048{\times}1024$ \\
        \midrule
        nanoVLM (\Nano)
        & 1{,}088
        & 576
        & 576 \\
        SmolVLM2
        & 1{,}088
        & 576
        & 576 \\
        LFM2.5-VL
        & 256
        & 242
        & 2{,}290 \\
        Qwen3.5-0.8B
        & 256
        & 512
        & 2{,}048 \\
        \textbf{\Flash}
        & \textbf{64}
        & \textbf{192}
        & \textbf{576} \\
        \bottomrule
    \end{tabular}
    \caption{
        Projected visual-token count across input resolutions.
    }
    \label{tab:visual_tokens}
\end{table}

Table~\ref{tab:visual_tokens} highlights the inefficiency of traditional preprocessing: paradoxically, baseline models like nanoVLM and SmolVLM2 generate nearly twice as many tokens for a small $512^2$ image (1,088 tokens) than for a large $2048{\times}1024$ image (576 tokens) due to forced isotropic upsampling. 
Conversely, at $512^2$, \Flash requires only the 64-token global view. This scales the visual prefix down by a massive factor of $17\times$ relative to \Nano and nanoVLM, and $4\times$ relative to Qwen3.5. At higher resolutions like $2048{\times}1024$, \Nano and \Flash predictably converge to the same 576-token layout, as the native input already requires no forced upscaling.

\subsection{Per-Backend TTFT Profiling}
\label{app:backend_results}

The main paper aggregates the optimal execution lane for each model-device pair to reflect real-world deployment scheduling. Table~\ref{tab:ttft_backend_full} provides the complete CPU and GPU measurements at both evaluated resolutions, strictly adhering to the protocol defined in Sec.~\ref{app:on_device_protocol}.

\begin{table*}[!t]
    \centering
    \scriptsize
    \setlength{\tabcolsep}{2.7pt}
    \begin{tabular}{@{}llrrrrrrrr@{}}
        \toprule
        Resolution & Model
        & \multicolumn{2}{c}{Pixel 9}
        & \multicolumn{2}{c}{Galaxy S23}
        & \multicolumn{2}{c}{S25 Ultra}
        & \multicolumn{2}{c}{iPhone 15} \\
        \cmidrule(lr){3-4}
        \cmidrule(lr){5-6}
        \cmidrule(lr){7-8}
        \cmidrule(l){9-10}
        & & CPU & GPU & CPU & GPU & CPU & GPU & CPU & GPU \\
        \midrule

        \multirow{5}{*}{$512^2$}
        & nanoVLM (\Nano)
        & 153.0 & 138.3
        & 119.0 & 116.7
        & 67.0 & 58.8
        & 70.3 & 10.9 \\

        & SmolVLM2
        & 128.2 & 136.1
        & 135.5 & 113.7
        & 48.7 & 64.3
        & 72.1 & 10.7 \\

        & LFM2.5-VL
        & 14.9 & 21.9
        & 13.6 & 14.2
        & 7.7 & 5.7
        & 3.4 & 0.4 \\

        & Qwen3.5-0.8B
        & 21.4 & 25.7
        & 22.0 & 19.9
        & 8.4 & 10.4
        & 5.0 & 0.7 \\

        & \textbf{\Flash}
        & \textbf{6.1} & \textbf{16.4}
        & \textbf{6.1} & \textbf{5.9}
        & \textbf{2.7} & \textbf{2.6}
        & \textbf{5.9} & \textbf{0.3} \\

        \midrule

        \multirow{5}{*}{$2048{\times}1024$}
        & nanoVLM (\Nano)
        & 70.3 & 70.3
        & 60.9 & 53.4
        & 33.2 & 36.6
        & 34.2 & 5.8 \\

        & SmolVLM2
        & 62.1 & 72.5
        & 70.9 & 59.9
        & 29.2 & 33.8
        & 36.7 & 5.4 \\

        & LFM2.5-VL
        & 140.3 & 103.8
        & 98.0 & 91.4
        & 70.3 & 43.0
        & 38.5 & 4.1 \\

        & Qwen3.5-0.8B
        & 311.4 & 325.2
        & 265.2 & 248.9
        & 127.5 & 137.3
        & 64.8 & 14.8 \\

        & \textbf{\Flash}
        & \textbf{40.9} & \textbf{58.8}
        & \textbf{45.2} & \textbf{39.6}
        & \textbf{19.3} & \textbf{22.6}
        & \textbf{30.7} & \textbf{2.6} \\
        \bottomrule
    \end{tabular}
    \caption{
        Warm TTFT (s) for CPU and GPU execution at $512^2$ and $2048{\times}1024$. Lower is better.
    }
    \label{tab:ttft_backend_full}
\end{table*}

\paragraph{Hardware dynamics and backend preference.}
As briefly noted in ~\cref{sec:on_device}, Table~\ref{tab:ttft_backend_full} reveals significant backend volatility across ecosystems. While GPU execution (Metal) systematically dominates on the iPhone due to Apple's mature unified memory architecture, Android devices frequently exhibit superior CPU performance for specific VLMs. Because TTFT encompasses multiple sequential stages (vision encoding, projection, and prefill), the theoretical GPU compute advantage on short visual prefixes is often bottlenecked by kernel launch latencies and synchronization overheads in Vulkan and OpenCL. Furthermore, on devices utilizing Mali GPUs (such as the Pixel 9), the quantized matrix multiplication kernels are generally weaker compared to the highly tuned ARM CPU execution path within \texttt{llama.cpp}. Consequently, the CPU matches or outperforms GPU TTFT on several Android platforms (notably the Pixel 9 and frequently the S25 Ultra), whereas the Adreno OpenCL backend typically maintains a performance edge on the Galaxy S23. Reporting the optimal lane dynamically therefore mirrors standard edge-deployment software routing.

\paragraph{TTFT speedups.}
Crucially, \Flash establishes the lowest TTFT on all four devices at both input resolutions under best-lane execution. At $2048{\times}1024$, it yields $1.35\times$ to $2.23\times$ speedups over nanoVLM, $1.51\times$ to $2.08\times$ over SmolVLM2, and $1.58\times$ to $2.54\times$ over LFM2.5-VL. As anticipated, the speedup gap relative to long-prefix baselines narrows at this high resolution compared to the extreme reductions seen at $512^2$. This precisely tracks the mathematical convergence of their visual-token budgets (~\cref{sec:on_device}), with any residual latency variations driven solely by intrinsic architectural differences and backend-specific kernel efficiencies.

\subsection{Peak Memory}
\label{app:peak_memory}

\begin{table*}[t]
    \centering
    \scriptsize
    \setlength{\tabcolsep}{2.7pt}
    \begin{tabular}{@{}llrrrrrrrr@{}}
        \toprule
        Resolution & Model
        & \multicolumn{2}{c}{Pixel 9}
        & \multicolumn{2}{c}{Galaxy S23}
        & \multicolumn{2}{c}{S25 Ultra}
        & \multicolumn{2}{c}{iPhone 15} \\
        \cmidrule(lr){3-4}
        \cmidrule(lr){5-6}
        \cmidrule(lr){7-8}
        \cmidrule(l){9-10}
        & & CPU & GPU & CPU & GPU & CPU & GPU & CPU & GPU \\
        \midrule

        \multirow{5}{*}{$512^2$}
        & nanoVLM (\Nano)
        & 1{,}236 & 2{,}224
        & 1{,}232 & 1{,}533
        & 1{,}251 & 1{,}552
        & 827 & 482 \\

        & SmolVLM2
        & 1{,}236 & 2{,}224
        & 1{,}232 & 1{,}534
        & 1{,}252 & 1{,}552
        & 780 & 482 \\

        & LFM2.5-VL
        & 501 & 1{,}123
        & 497 & 588
        & 510 & 602
        & 238 & 181 \\

        & Qwen3.5-0.8B
        & 866 & 2{,}243
        & 853 & 942
        & 878 & 972
        & 903 & 836 \\

        & \textbf{\Flash}
        & 750 & 1{,}692
        & 746 & 1{,}047
        & 763 & 1{,}066
        & 292 & 285 \\

        \midrule

        \multirow{5}{*}{$2048{\times}1024$}
        & nanoVLM (\Nano)
        & 992 & 1{,}949
        & 988 & 1{,}290
        & 1{,}005 & 1{,}305
        & 754 & 420 \\

        & SmolVLM2
        & 993 & 1{,}939
        & 989 & 1{,}290
        & 1{,}005 & 1{,}306
        & 719 & 420 \\

        & LFM2.5-VL
        & 535 & 1{,}197
        & 530 & 621
        & 543 & 636
        & 271 & 216 \\

        & Qwen3.5-0.8B
        & 1{,}080 & 2{,}495
        & 1{,}066 & 1{,}160
        & 1{,}081 & 1{,}174
        & 1{,}153 & 941 \\

        & \textbf{\Flash}
        & 964 & 1{,}915
        & 960 & 1{,}261
        & 979 & 1{,}279
        & 567 & 399 \\

        \bottomrule
    \end{tabular}
    \caption{
        Peak resident memory (MiB) during on-device inference at $512^2$ and $2048{\times}1024$. Measurements exclude model loading and follow the protocol in Sec.~\ref{app:on_device_protocol}.
    }
    \label{tab:peak_memory_full}
\end{table*}

We measure the peak process-resident memory footprint during the complete multimodal inference execution, strictly excluding model loading overhead. We periodically sample the resident memory during execution. For each repetition, we record the maximum observed usage, and we report the median peak across runs. Values are reported in MiB for both CPU and GPU execution paths. Because Android and iOS employ fundamentally different operating system memory-accounting conventions, memory measurements are interpreted strictly within each hardware platform rather than across ecosystems.

At $512^2$, the drastically reduced visual prefix of \Flash yields a substantial reduction in runtime memory relative to full-resolution baselines (nanoVLM and SmolVLM2). Across Android CPU and GPU execution paths, \Flash lowers peak memory by $24\%$ to $39\%$, whereas on iPhone 15, the reduction reaches $41\%$ to $65\%$. Furthermore, \Flash maintains a lower memory usage than Qwen3.5 across six out of eight device-backend configurations at this resolution.

At $2048{\times}1024$, the memory advantage relative to full-resolution baselines narrows as visual-token allocations converge. Nevertheless, \Flash remains below Qwen3.5 in six out of eight device-backend pairs. Jointly with the TTFT profiling (Sec.~\ref{app:backend_results}), these measurements demonstrate that scaling down the visual sequence simultaneously lowers multimodal prefill latency and peak memory usage. Finally, the consistently higher GPU memory footprint observed on Android stems from Vulkan/OpenCL driver overheads and explicit workspace buffer allocations, whereas Apple's Metal backend natively exploits zero-copy unified memory to maintain near-parity between execution lanes.

\subsection{Autoregressive Decode Throughput}
\label{app:decode_throughput}

We measure autoregressive decode throughput independently of the multimodal prefill stage, explicitly excluding vision encoding, multimodal projection, and language-model prefill overheads. Throughput is computed from the number of generated tokens and the corresponding decode wall time, reported in tokens per second. We adhere to the fixed-length protocol defined in Sec.~\ref{app:on_device_protocol}, which forces 32 output tokens and disables early EOS termination to ensure strictly comparable measurements across models. Reported values are medians over the warm repetitions.

Across both resolutions, \Flash achieves higher decode throughput than nanoVLM and SmolVLM2 in 13 out of 16 device-backend configurations and outperforms Qwen3.5 in 15 out of 16. At $512^2$, when selecting the faster execution lane on each device, \Flash provides $1.1\times$ to $1.8\times$ higher throughput than nanoVLM and SmolVLM2, and $1.3\times$ to $2.1\times$ higher throughput than Qwen3.5.

At $2048{\times}1024$, \Flash remains faster than nanoVLM and SmolVLM2 on three of the four devices under best-lane execution, and exceeds Qwen3.5 by $1.1\times$ to $2.9\times$ across all four. While LFM2.5-VL achieves the highest absolute decode throughput across most configurations due to its highly optimized, lightweight language-model backbone, \Flash remains highly competitive. Together with the TTFT results, these measurements confirm that the dramatic reduction in multimodal prefill cost does not introduce any corresponding penalty during the autoregressive generation phase. Absolute decode throughput remains inherently dependent on the underlying language-model architecture and execution backend, while \Flash consistently maintains strong generation performance across the evaluated hardware spectrum.

\begin{table*}[ht]
    \centering
    \scriptsize
    \setlength{\tabcolsep}{2.7pt}
    \begin{tabular}{@{}llrrrrrrrr@{}}
        \toprule
        Resolution & Model
        & \multicolumn{2}{c}{Pixel 9}
        & \multicolumn{2}{c}{Galaxy S23}
        & \multicolumn{2}{c}{S25 Ultra}
        & \multicolumn{2}{c}{iPhone 15} \\
        \cmidrule(lr){3-4}
        \cmidrule(lr){5-6}
        \cmidrule(lr){7-8}
        \cmidrule(l){9-10}
        & & CPU & GPU & CPU & GPU & CPU & GPU & CPU & GPU \\
        \midrule

        \multirow{5}{*}{$512^2$}
        & nanoVLM (\Nano)
        & 15.6 & 27.9
        & 23.2 & 17.8
        & 47.0 & 51.9
        & 8.3 & 52.6 \\

        & SmolVLM2
        & 18.2 & 28.1
        & 18.5 & 18.3
        & 64.8 & 49.6
        & 11.4 & 53.6 \\

        & LFM2.5-VL
        & 42.6 & 47.8
        & 52.3 & 42.6
        & 69.8 & 122.4
        & 28.2 & 116.0 \\

        & Qwen3.5-0.8B
        & 17.9 & 22.9
        & 12.8 & 21.1
        & 45.2 & 24.6
        & 6.9 & 40.5 \\

        & \textbf{\Flash}
        & 28.1 & 30.7
        & 32.7 & 33.8
        & 93.5 & 93.7
        & 7.6 & 78.9 \\

        \midrule

        \multirow{5}{*}{$2048{\times}1024$}
        & nanoVLM (\Nano)
        & 17.4 & 31.4
        & 25.7 & 20.5
        & 56.5 & 63.2
        & 10.9 & 53.6 \\

        & SmolVLM2
        & 22.4 & 32.8
        & 20.7 & 22.1
        & 71.8 & 55.0
        & 13.6 & 54.3 \\

        & LFM2.5-VL
        & 26.3 & 48.9
        & 38.5 & 35.0
        & 39.7 & 76.3
        & 28.9 & 123.4 \\

        & Qwen3.5-0.8B
        & 12.6 & 20.6
        & 15.3 & 17.0
        & 36.8 & 32.5
        & 7.0 & 24.4 \\

        & \textbf{\Flash}
        & 23.3 & 13.3
        & 27.7 & 23.3
        & 89.7 & 66.2
        & 9.8 & 71.7 \\
        \bottomrule
    \end{tabular}
    \caption{
        Autoregressive decode throughput (tokens/s) at $512^2$ and $2048{\times}1024$.
    }
    \label{tab:decode_throughput_full}
\end{table*}

\subsection{End-to-End Latency}
\label{app:ttlt}

We complement the fixed-length benchmarks with time to last token (TTLT) to evaluate end-to-end responsiveness under free-form generation. TTLT captures the complete inference pipeline (vision encoding, multimodal projection, language-model prefill, and autoregressive decoding) while strictly excluding model load times. Using the prompt \textit{``Please describe the image in one sentence.''}, we evaluate inputs at both $512^2$ and $2048{\times}1024$ resolutions. Models decode naturally until the end-of-sequence token, subject to a 4{,}096-token safety cap. To rigorously account for runtime variance and model-specific verbosity, we report the median TTLT over warm repetitions alongside the median number of generated tokens.

\begin{figure}[t]
    \centering
    \includegraphics[width=\linewidth]{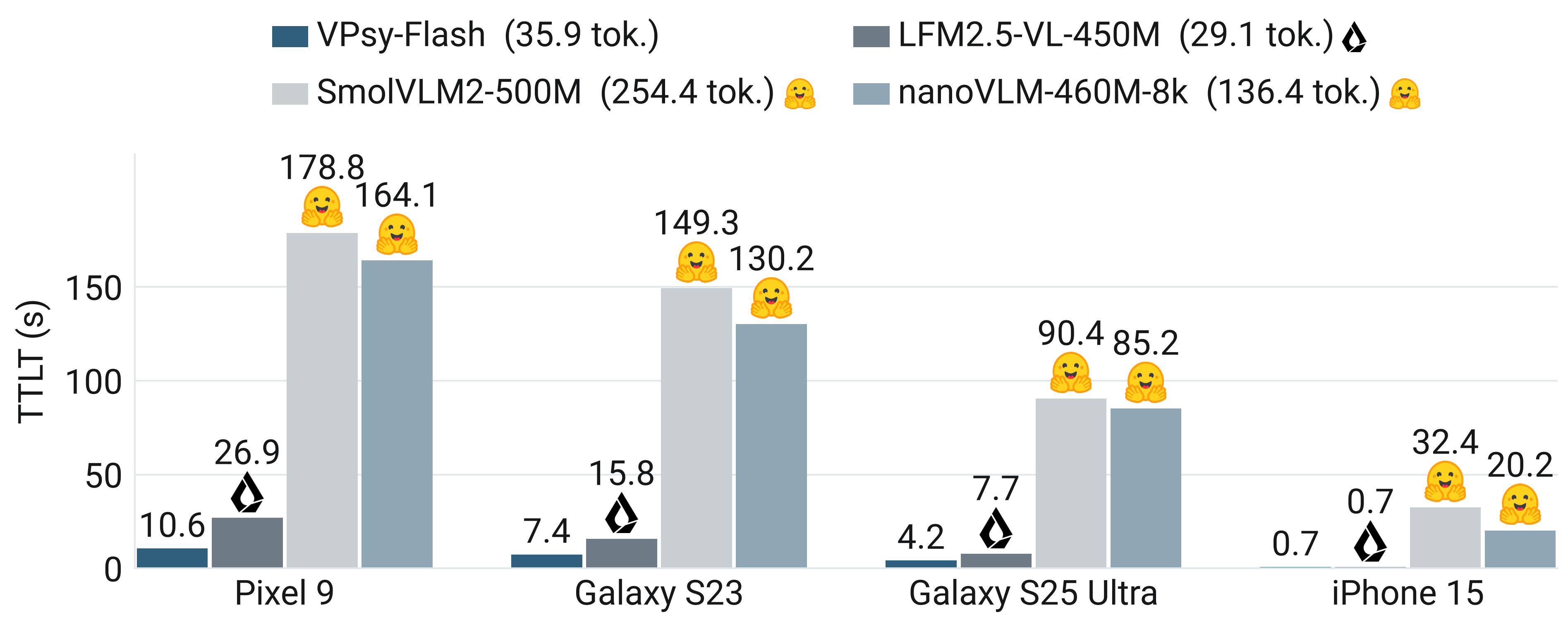}
    \caption{
        End-to-end time to last token for free-form generation. The optimal CPU or GPU execution lane is reported for each device.
    }
    \label{fig:ttlt}
\end{figure}

As shown in Fig.~\ref{fig:ttlt}, \Flash achieves the lowest TTLT on all three Android devices, completing generation in 10.6\,s on the Pixel~9, 7.4\,s on the Galaxy~S23, and 4.2\,s on the Galaxy~S25 Ultra. Relative to LFM2.5-VL-450M, this corresponds to $2.5\times$, $2.1\times$, and $1.8\times$ lower end-to-end latency, respectively; on the iPhone~15, both models complete generation in a highly interactive 0.7\,s.

Crucially, this latency advantage is not an artifact of shorter responses. \Flash generates a median of 35.9 tokens, compared with 29.1 for LFM2.5-VL-450M. Furthermore, nanoVLM and SmolVLM2 generate substantially longer responses to the same instruction, averaging 136.4 and 254.4 tokens, respectively. By capturing both raw execution efficiency and the computational cost of architectural verbosity, TTLT confirms the core premise of our approach: for practical on-device workloads, the multimodal prefill dictates responsiveness. \Flash retains its massive efficiency advantage precisely because its compressed visual footprint eliminates the primary latency bottleneck before generation even begins.
\section{Additional Ablations}
\label{app:additional_ablations}

\subsection{Robustness to the choice of judge.}
\label{app:judge_robustness}

Because LLM-based evaluation introduces an additional evaluator choice into the scoring pipeline, we verify that our main comparison is not specific to the judge used in the primary evaluation. We re-score the judge-dependent benchmarks for \Nano and the strongest competing model, LFM2.5-VL-450M, using GPT-4o-mini and GPT-5-chat in addition to Qwen3.6-27B, while keeping the remaining rule-scored benchmarks unchanged.

As shown in Table~\ref{tab:judge-robustness}, absolute scores vary across judges, particularly on benchmarks such as MM-IFEval and MMVet. However, the overall comparison remains highly stable. \Nano outperforms LFM2.5-VL-450M on 16 of 17 benchmarks under all three judges, while the normalized aggregate margin remains nearly unchanged at $+2.7$, $+2.9$, and $+2.8$ points under Qwen3.6-27B, GPT-4o-mini, and GPT-5-chat, respectively. These results indicate that the reported performance advantage is robust to the choice of LLM judge rather than an artifact of a particular evaluator.

\begin{table}[tb]
  \centering
  \footnotesize
  \setlength{\tabcolsep}{3pt}
  \begin{tabular}{@{}l cc cc cc@{}}
    \toprule
    & \multicolumn{2}{c}{Qwen3.6-27B} & \multicolumn{2}{c}{GPT-4o-mini}
    & \multicolumn{2}{c}{GPT-5-chat}\\
    \cmidrule(lr){2-3}\cmidrule(lr){4-5}\cmidrule(lr){6-7}
    Benchmark & Ours & LFM & Ours & LFM & Ours & LFM\\
    \midrule
    OCRBench   & \textbf{757}  & 710  & \textbf{775}  & 725  & \textbf{766}  & 711\\
    DocVQA     & \textbf{85.7} & 82.7 & \textbf{85.1} & 82.4 & \textbf{85.7} & 82.7\\
    ChartQA    & \textbf{78.7} & 76.6 & \textbf{78.2} & 76.7 & \textbf{78.8} & 77.0\\
    InfoVQA    & \textbf{49.8} & 48.9 & \textbf{50.0} & 49.0 & \textbf{50.0} & 49.0\\
    TextVQA    & \textbf{79.3} & 78.8 & \textbf{78.2} & 77.1 & \textbf{80.2} & 79.5\\
    MathVista  & \textbf{48.9} & 42.2 & \textbf{48.9} & 42.3 & \textbf{49.0} & 42.3\\
    MM-IFEval  & \textbf{42.3} & 42.0 & 36.6 & \textbf{36.9} & \textbf{49.4} & 47.8\\
    MMVet      & 32.3 & \textbf{36.0} & \textbf{27.0} & 26.3 & 33.3 & \textbf{37.6}\\
    \midrule
    Avg$_{\text{norm}}$ (17) & \textbf{62.3} & 59.6 & \textbf{61.6} & 58.7 & \textbf{62.9} & 60.1\\
    \bottomrule
  \end{tabular}
  \caption{Robustness to the choice of LLM judge. ``Ours'' is \Nano and ``LFM''
    is LFM2.5-VL-450M, the previous best competitor in the $\sim$0.5B cohort. Both are
    rescored under three judges on the eight benchmarks whose scoring depends on
    the judge; the remaining nine are multiple-choice or rule-scored and return
    identical values under all three, so they are omitted here and included
    unchanged in Avg$_{\text{norm}}$. Absolute scores shift with the judge, most
    visibly on MM-IFEval, but the comparison does not: Ours leads by 2.7, 2.9 and
    2.8 normalized points under Qwen3.6-27B, GPT-4o-mini and GPT-5-chat, and
    takes 16 of 17 benchmarks under every judge. Higher score per judge in
    \textbf{bold}.}
  \label{tab:judge-robustness}
\end{table}

Judge scoring also does not favor our models relative to the native
metrics. Among the 17 benchmarks, the 5 benchmarks marked $\dagger$ in Tab.~1 (OCRBench, DocVQA, ChartQA, InfoVQA, TextVQA) are the only ones where our protocol replaces the conventional metric with a
judge; the remaining 12 are scored exactly as convention dictates,
nine by multiple-choice or rule-based matching with no judge
involvement, and three (MathVista, MM-IFEval, MMVet) by LLM-assisted
grading that is already part of their official protocols. The
strict-versus-judge comparison therefore applies to precisely the five
benchmarks where our protocol deviates from convention.
Table~\ref{tab:judge-delta} reports it across the cohort. The correction
is universal, the judge rescues every model on all five benchmarks, and
its largest beneficiaries are the baselines rather than our checkpoints:
SmolVLM2-500M and the base checkpoint gain 6.6 and 6.7 points on average
against 5.0 for our two models, consistent with their far more verbose
responses.

\begin{table}[tb]
\centering
\footnotesize
\setlength{\tabcolsep}{3pt}
\resizebox{.8\linewidth}{!}{%
\begin{tabular}{@{}l ccc cc@{}}
  \toprule
  & \multicolumn{3}{c}{Baselines} & \multicolumn{2}{c}{Ours}\\
  \cmidrule(lr){2-4}\cmidrule(lr){5-6}
  Benchmark & nanoVLM & SmolVLM2 & LFM & \Nano & \Flash\\
  \midrule
  ChartQA   & +4.5  & +4.0  & +1.6 & +1.8 & +2.3\\
  DocVQA    & +5.0  & +6.1  & +4.4 & +5.1 & +4.4\\
  InfoVQA   & +11.0 & +11.4 & +5.6 & +6.5 & +6.3\\
  TextVQA   & +9.6  & +10.7 & +8.9 & +8.4 & +8.1\\
  OCRBench  & +35   & +8    & +31  & +31  & +39\\
  \midrule
  Avg       & 6.7   & 6.6   & 4.7  & 5.0  & 5.0\\
  \bottomrule
\end{tabular}}
\caption{Score delta (judge $-$ strict) each model receives on the
five benchmarks where our evaluation protocol replaces the conventional strict
metric with a judge; OCRBench is on its native /1000 scale and divided
by 10 in the Avg row. The judge rescues every model on every benchmark,
and the largest corrections go to the baselines rather than to our
models.}
\label{tab:judge-delta}
\end{table}


\subsection{Impact of Post-Training under the Bounded Native-Resolution Policy}
\label{app:sft_ablation}

In the main text (Sec.~4), we argued that enforcing a strictly bounded visual-token budget solely at inference time induces a severe distribution shift, making post-training essential. To empirically validate this design choice, Table~\ref{tab:flash_sft_ablation} presents an ablation study comparing a zero-shot application of our visual-token policy against the fully adapted \Flash checkpoint. For the \textbf{Zero-shot} baseline, we take the \Nano and force it to evaluate the benchmark suite using the \Flash preprocessing logic. For the \textbf{Full Recipe} configuration, we evaluate our final \Flash model, which has been completely adapted under this compressed visual budget.

\begin{table}[htbp]
    \centering
    \footnotesize
    \setlength{\tabcolsep}{5pt}
    \begin{tabular}{@{}lccc@{}}
        \toprule
        \textbf{Benchmark}
        & \textbf{Zero-shot}
        & \textbf{Full Recipe}
        & \textbf{$\Delta$} \\
        \midrule

        \multicolumn{4}{@{}l}{\textit{Document Understanding \& OCR}} \\
        \midrule
        OCRBench    & 709  & 738  & +29 \\
        DocVQA      & 85.4 & 85.1 & -0.3 \\
        ChartQA     & 69.4 & 77.4 & +8.0 \\
        InfoVQA     & 48.8 & 51.1 & +2.3 \\
        TextVQA     & 73.3 & 75.6 & +2.3 \\

        \midrule
        \multicolumn{4}{@{}l}{\textit{Visual Perception}} \\
        \midrule
        MME         & 1540 & 1619 & +79 \\
        SEED        & 66.4 & 67.6 & +1.2 \\
        MMBench     & 60.6 & 60.5 & -0.1 \\
        RealWorldQA & 58.8 & 58.4 & -0.4 \\

        \midrule
        \multicolumn{4}{@{}l}{\textit{Reasoning \& Knowledge}} \\
        \midrule
        ScienceQA   & 83.7 & 84.0 & +0.3 \\
        AI2D        & 66.2 & 66.4 & +0.2 \\
        MMStar      & 46.1 & 45.5 & -0.6 \\
        MMMU        & 33.3 & 30.7 & -2.6 \\
        MathVista   & 44.4 & 47.8 & +3.4 \\
        MMVet      & 32.9 & 31.1 & -1.8 \\

        \midrule
        \multicolumn{4}{@{}l}{\textit{Instruction Following \& Reliability}} \\
        \midrule
        MM-IF       & 42.5 & 43.7 & +1.2 \\
        POPE        & 83.0 & 87.5 & +4.5 \\

        \midrule
        \textbf{Avg\textsubscript{norm}}
        & \textbf{60.1}
        & \textbf{61.4}
        & \textbf{+1.3} \\
        \bottomrule
    \end{tabular}

    \vspace{2mm}
    \caption{Impact of our complete post-training recipe under the bounded native-resolution policy. The \textit{Zero-shot} baseline evaluates the full-resolution checkpoint using the \Flash preprocessing logic strictly at inference time, whereas \textit{Full Recipe} denotes our final \Flash model.}
    \label{tab:flash_sft_ablation}
\end{table}

The empirical results confirm that reducing the visual-token budget strictly at inference time severely degrades the model's representational capacity, resulting in a baseline normalized score of 60.1. By executing our complete post-training recipe under the restricted prefix (Full Recipe), we successfully recover and enhance performance, achieving an overall normalized average of 61.4. Without this comprehensive adaptation, the model struggles to parse fine-grained details packed within a drastically smaller token grid; following our proposed pipeline, we observe substantial recoveries in ChartQA (+8.0 points), OCRBench (+29 points), and MME (+79 points). Furthermore, the complete adaptation significantly mitigates object hallucination triggered by the sudden reduction in visual context, as reflected by a +4.5 point gain on POPE. While a few reasoning benchmarks exhibit minor fluctuations (e.g., MMMU, MMVet), the overall normalized average demonstrates that our full post-training recipe is strictly necessary to restore and stabilize semantic capacity when deploying aggressively compressed visual prefixes.
\section{Implementation Details}
\label{app:implementation_details}

\subsection{Evaluation Protocol}
\label{app:eval_protocol}

\paragraph{Harness and metrics.}
All models are scored in-house under a single VLMEvalKit~\cite{duan2024vlmevalkit}
harness, so every entry in the main paper shares the same prompts, harness
configuration and metric implementations. We report each benchmark's official
metric: accuracy for most, F1 for POPE, partial credit for MMVet,
instruction-following accuracy for MM-IFEval, perception plus reasoning points
for MME, and the $/1000$ score for OCRBench.

\paragraph{Normalization.}
Avg$_{\text{norm}}$ is the unweighted mean of the \num{17} per-benchmark scores
after linear normalization to $0$--$100$: MME is divided by $2800$, OCRBench by
$1000$, and the remaining benchmarks are already percentages and enter unchanged.
No benchmark is weighted, dropped or capped. The same constants are used for
every table in the paper, including the stage-wise and merging ablations, so
aggregate scores are comparable across all of them.

\paragraph{LLM-as-judge scoring.}
OCRBench, DocVQA, ChartQA, InfoVQA and TextVQA accept free-form answers and are
conventionally scored by exact string match. At the $\sim$0.5B scale that
convention conflates reading accuracy with output-format compliance, because a
correct answer is marked wrong when it differs from the reference only in surface
form, such as ``12\%'' against ``12 percent'', a paraphrase, a unit, or a
formatting difference. Models in this cohort vary widely in how tightly they
follow answer-format instructions, so exact match systematically understates the
reading accuracy of models that answer verbosely. We therefore score these five
with an LLM judge (Qwen3.6-27B~\cite{qwen2026qwen36}) and mark them $\dagger$
throughout. Because the judge is itself a modeling choice,
\cref{tab:judge-robustness} summarizes the headline comparison under three
independent judges; absolute scores shift, and the ordering and aggregate margin do not.

\subsection{Training Details}
In this section, we present the training details for each stage. The datasets of each stage are summarized in Table \ref{tab:dataset-stats}.

\label{app:stage1_ID}
For the stage 1, we fine-tune nanoVLM-460M from the public nanoVLM-460M-8k checkpoint on a benchmark decontaminated Nemotron training split of \(5{,}950{,}891\) image--question--answer triples.

We guard against train--test leakage before Stage~1. For every training and evaluation image we compute a 64-bit DCT perceptual hash (pHash) and a 64-bit difference hash (dHash); a training image is flagged when it lies within Hamming distance~5 of any evaluation image (exact or near-duplicate), using a banded index for recall. All flagged images and their question--answer rows are removed (Tab.~\ref{tab:decontam}).

\begin{table}[t]
\centering\small
\begin{tabular}{lr}
\toprule
Benchmark & Overlapping images \\
\midrule
SEEDBench & 4{,}463 \\ ChartQA & 1{,}611 \\ ScienceQA & 1{,}610 \\ POPE & 1{,}142 \\
AI2D & 631 \\ DocVQA & 560 \\ MMStar & 459 \\ TextVQA & 407 \\ MathVista & 358 \\
MMBench & 277 \\ OCRBench & 175 \\ MMMU & 175 \\ MM-IF & 94 \\ MME & 84 \\
InfoVQA & 49 \\ RealWorldQA & 9 \\ MMVet & 0 \\
\bottomrule
\end{tabular}
\caption{Per-benchmark count of held-out evaluation images that appear in
(and are removed from) the raw training corpus. 
Counts are non-exclusive overlap
events: one training image may overlap multiple benchmarks.
In total 5{,}741
unique training images are removed; since one image is typically reused across
several question--answer pairs, this corresponds to 35{,}649 training samples.}
\label{tab:decontam} 
\end{table}
Multi-page samples are resolved offline with a page-selection heuristic that retains a single aligned image when the question uniquely identifies a page, and all chain-of-thought segments are stripped so that supervision targets short final answers only.
Training uses distributed data-parallel supervised fine-tuning for four epochs in \texttt{bfloat16} with a maximum sequence length of \(4{,}096\) tokens, a per-GPU batch size of \(1\), gradient accumulation of \(8\), and \(80\) GPUs (\(10{\times}8\)), yielding an effective batch size of \(640\) (approximately \(37\)k optimizer updates).
Learning rates are set to \(1{\times}10^{-4}\) for the modality projector, \(5{\times}10^{-5}\) for the language model, and \(1{\times}10^{-5}\) for the vision encoder.

For Stage 2, we initialize from the final Stage-1 checkpoint and perform continued SFT on a mixed dataset combining the benchmark-decontaminated Nemotron VQA data with FineVision and MM-IFEngine instruction data under a decoded merge. 
Training uses the same distributed data-parallel setup as Stage 1, with \texttt{bfloat16}, a maximum sequence length of \(4{,}096\) tokens, and a long-side image resolution of \(2{,}048\) pixels, but on \(40\) GPUs with a per-GPU batch size of \(1\) and gradient accumulation of \(8\), yielding an effective batch size of \(320\). 
Since training is warm-started from the strong Stage 1 checkpoint, we reduce the learning rates to \(5{\times}10^{-5}\), \(2.5{\times}10^{-5}\), and \(5{\times}10^{-6}\) for the modality projector, language model, and vision encoder, respectively. 



For Stage~3, we initialize from the Stage-2 checkpoint and continue SFT with InfoSFT.
To prevent catastrophic forgetting we replay \(2\)M examples from the Stage-2 mix (UPMB2M).
We further add \(1{,}730{,}095\) non-replay examples from the Stage-3 sources listed in Table~\ref{tab:dataset-stats}.
Optimization uses the information-aware token-weighted InfoSFT loss with confidence threshold \(P{=}0.87\).
The per-token weight is \(w(p)\propto p\cdot\operatorname{ReLU}\!\bigl(\operatorname{logit}(P)-\operatorname{logit}(p)\bigr)\)~\cite{sabbaghi2026infosft}, which is zero for already-confident tokens (\(p\geq P\)) and emphasizes medium-confidence tokens.
Training runs for two epochs in \texttt{bfloat16} with a maximum sequence length of \(8{,}192\) tokens and long-side image resize to \(2{,}048\) pixels, on \(16\) GPUs (\(2{\times}8\)) with per-GPU batch size \(2\) and gradient accumulation of \(8\) (effective batch size \(256\)).
Learning rates are \(5{\times}10^{-6}\) for the modality projector and language model and \(2{\times}10^{-6}\) for the vision encoder, with AdamW, cosine decay after a \(3\%\) warmup, weight decay \(0.01\), and gradient clipping at \(1.0\).

In Stage~4 we merge six checkpoints from these specialists that share the Stage-2 initialization, using TIES-Merging~\cite{yadav2023ties}.
Each checkpoint is encoded as a task vector \(\tau_t=\theta_t-\theta_2\); vectors are magnitude-trimmed to their top \(k{=}20\%\) entries; a mass-weighted majority vote elects a sign per coordinate (\(\operatorname{sign}\) of the sum of trimmed values); and only sign-agreeing parameters are averaged (disjoint mean).
The merged task vector is added back as \(\theta\leftarrow\theta_2+\lambda\tau\), with \(\lambda\) selected as described in Sec.~3.

\begin{table*}
\centering
\small
\setlength{\tabcolsep}{4pt}
\begin{tabular}{cllr}
\toprule
\textbf{Stage} & \textbf{Corpus} & \textbf{Description} & \textbf{\#Examples} \\
\midrule
1 & Filtered Nemotron-V3{~\cite{nvidia2025nemotron}}
  & Decontam.\ + Single-image filtering; think tokens stripped
  & 5{,}950{,}891 \\
\midrule
\multirow{2}{*}{2}
  & Upperbound Mixed (full)
  & Filtered Nemotron-V3{~\cite{nvidia2025nemotron}} + Sampled FineVision{~\cite{finevision2025}} + {MM-IFEngine~\cite{mmif23k_iccv25}} 
  & 7{,}527{,}286 \\
  & UPMB2M
  & Capability-balanced 2M subset of upperbound Mixed
  & 2{,}000{,}000 \\
\midrule
\multirow{10}{*}{3}
  & 3Bsynth 
  & Synthetic / full-real preference-style mix
  & 656{,}160 \\
  & MathEnrich (total)
  & Decontaminated math/chart/geometry SFT
  & 448{,}088 \\
  & \quad MathV360K (filtered){~\cite{shi2024mathllava}}
  & Kept math/chart/science sources only
  & 176{,}879 \\
  & \quad Geo170K{~\cite{gao2025gllava}}
  & Geometry QA / reasoning
  & 115{,}814 \\
  & \quad TinyChart-PoT{~\cite{zhang2024tinychart}}
  & Chart QA with program-of-thought answers
  & 155{,}395 \\
  & OCR/Doc enrichment 
  & Docmatix{~\cite{laurencon2024building}} + OCR + TextVQA{~\cite{singh2019textvqa}} refresh (net add-on)
  & 282{,}240 \\
  & ChartQA{~\cite{masry2022chartqa}} 
  & Chart question answering
  & 17{,}742 \\
  & Chart2Text{~\cite{kantharaj2022charttotext}} 
  & Chart captioning / description
  & 28{,}972 \\
  & CoSyn-400K{~\cite{yang2025cosyn}} 
  & Text-rich synthetic doc/table/chart/math QA
  & 296{,}893 \\
  & + UPMB2M
  & Capability-balanced 2M subset of upperbound Mixed
  & 2{,}000{,}000 \\
\midrule
\multirow{2}{*}{5}
  & Anti-doom-loop subsets
  & Anti-doom-loop self-mine mix {+} capability replay
  & 29{,}830 \\
  & Safety alignment subsets phase 1
  & Phase 1 high-conf.\ self-mined safety pairs {+} capability replay
  & 13{,}703 \\
  & Safety alignment subsets phase 2
  & Phase 2 high-conf.\ self-mined safety pairs {+} capability replay
  & 6{,}413 \\
\bottomrule
\end{tabular}
\caption{Training-corpus statistics for the datasets used in different stages.
Counts are example rows after decontamination / filtering where applicable.}
\label{tab:dataset-stats}
\end{table*}

In the final stage, we perform behavior alignment with an MPO~\cite{wang2024mpo} objective combining DPO, BCO, and chosen-response SFT.
Training proceeds sequentially through doom-loop stabilization and safety alignment.
We first train for \(1{,}600\) steps on \(29{,}830\) capability-balanced preference pairs containing same-prompt anti-doom-loop examples, OCR and instruction-following data, and reasoning replay.
We then perform two safety-alignment phases: \(200\) steps on \(13{,}703\) fixed safety and replay examples, followed by \(100\) steps on \(4{,}275\) self-mined safety pairs mixed with \(2{,}138\) capability-replay examples.
The MPO loss weights \((w_{\mathrm{DPO}}, w_{\mathrm{BCO}}, w_{\mathrm{SFT}})\) are \((0.8, 0.2, 0.5)\), \((0.4, 0.1, 1.0)\), and \((0.25, 0.05, 1.0)\) for the three phases, respectively, with \(\beta=0.5\).
All runs use \(8\) GPUs, \texttt{bfloat16}, a maximum sequence length of \(4{,}096\), long-side image resolution of \(2{,}048\) pixels, and an effective batch size of \(64\).
The vision encoder is frozen, while the modality-projector/language-model learning rates are progressively reduced from \(1{\times}10^{-5}/5{\times}10^{-6}\) to \(2{\times}10^{-6}/1{\times}10^{-6}\) and \(1{\times}10^{-6}/5{\times}10^{-7}\) across the three phases.
\section{Qualitative Results and On-Device Deployment}
\label{app:qualitative_appendix}

In this section, we provide qualitative generation examples to demonstrate the retention of core capabilities in our models, followed by a real-world deployment scenario that highlights the dramatic interactivity improvements achieved by the \Flash variant.

\subsection{Qualitative Comparisons}
\label{app:qualitative_results}
\begin{figure*}[htbp]
    \centering
    
    \begin{minipage}[c]{0.20\textwidth}
        \centering
        \includegraphics[width=\linewidth]{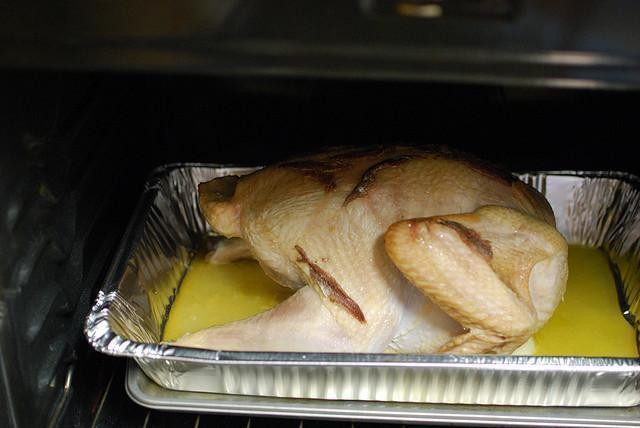}
    \end{minipage}
    \hfill
    \begin{minipage}[c]{0.78\textwidth}
        \textbf{Q:} Is the pan sitting directly on the oven rack? \\[0.3em]
        \textbf{\textit{\Nano}}: Yes. \cmark\\
        \textbf{\textit{\Flash}}: Yes, the pan is sitting directly on the oven rack. \cmark\\
        \textbf{\textit{nanoVLM}}: No. \xmark\\
        \textbf{\textit{LFM2.5-VL}}: No. \xmark
    \end{minipage}
    
    \vspace{0.4cm}
    \hrule
    \vspace{0.4cm}
    
    \begin{minipage}[c]{0.20\textwidth}
        \centering
        \includegraphics[width=\linewidth]{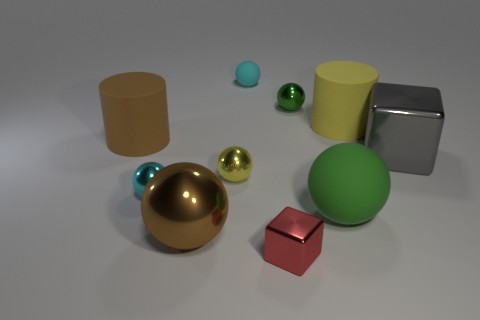}
    \end{minipage}
    \hfill
    \begin{minipage}[c]{0.78\textwidth}
        \textbf{Q:} Is the red cube to the left or right of the big green ball? \\[0.3em]
        \textbf{\textit{\Nano}}: Left. \cmark\\
        \textbf{\textit{\Flash}}: Left. \cmark\\
        \textbf{\textit{nanoVLM}}: Left. \cmark\\
        \textbf{\textit{LFM2.5-VL}}: The red cube is positioned to the right of the large green sphere. It's a small, shiny red cube that stands out against the other objects in the scene, which include various spheres and cylinders in different colors and sizes. \xmark
    \end{minipage}
    
    \vspace{0.4cm}
    \hrule
    \vspace{0.4cm}
    
    \begin{minipage}[c]{0.20\textwidth}
        \centering
        \includegraphics[width=\linewidth]{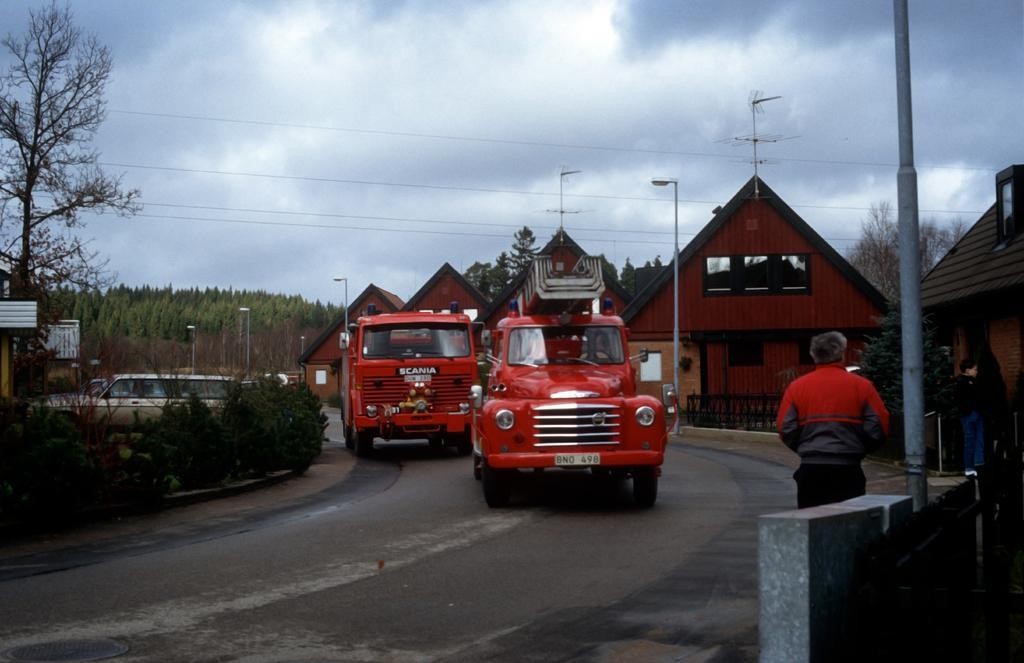}
    \end{minipage}
    \hfill
    \begin{minipage}[c]{0.78\textwidth}
        \textbf{Q:} Can you give me a brief description of what's happening in this photo? \\[0.3em]
        \textbf{\textit{\Nano}}: A red fire truck is driving down a street, with another fire truck following behind. There are a few people standing on the sidewalk, watching the vehicles. \cmark\\
        \textbf{\textit{\Flash}}: The image depicts a street scene with two red Scania trucks driving down a residential street. The trucks are positioned in the middle of the road, and the driver of the truck in the foreground is visible. The setting appears to be a suburban area with houses in the background, and the sky is overcast. \cmark\\
        \textbf{\textit{nanoVLM}}: In this image we can see vehicles on the road. We can also see a person standing on the right side. We can also see a pole, a fence, a building, trees, a pole, wires, a vehicle on the road, a person standing on the left side, a pole, a pole, a tree and the sky which looks cloudy. \xmark\\
        \textbf{\textit{LFM2.5-VL}}: In this image I can see few vehicles on the road. I can also see few people, poles, trees, houses, wires and the sky. \xmark
    \end{minipage}
    
    \vspace{0.4cm}
    \hrule
    \vspace{0.4cm}
    
    \begin{minipage}[c]{0.20\textwidth}
        \centering
        \includegraphics[width=\linewidth]{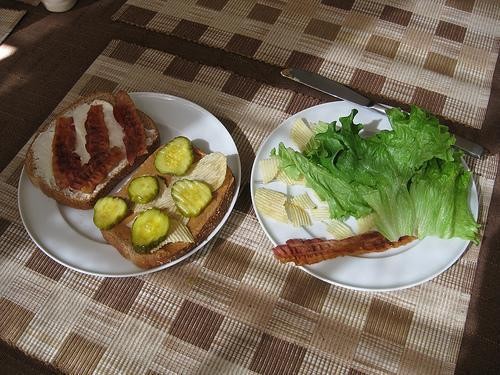}
    \end{minipage}
    \hfill
    \begin{minipage}[c]{0.78\textwidth}
        \textbf{Q:} How many pickles are on the sandwich? \\[0.3em]
        \textbf{\textit{\Nano}}: 5. \cmark\\
        \textbf{\textit{\Flash}}: 5. \cmark\\
        \textbf{\textit{nanoVLM}}: Six. \xmark\\
        \textbf{\textit{LFM2.5-VL}}: There are three pickles on the sandwich. They are placed on the left side of the sandwich, which is topped with a spread of what appears to be mayonnaise or a similar condiment. The pickles are arranged in a row, adding a pop of green color to the sandwich. \xmark
    \end{minipage}
    
    \vspace{0.4cm}
    \hrule
    \vspace{0.4cm}
    
    \begin{minipage}[c]{0.20\textwidth}
        \centering
        \includegraphics[width=\linewidth]{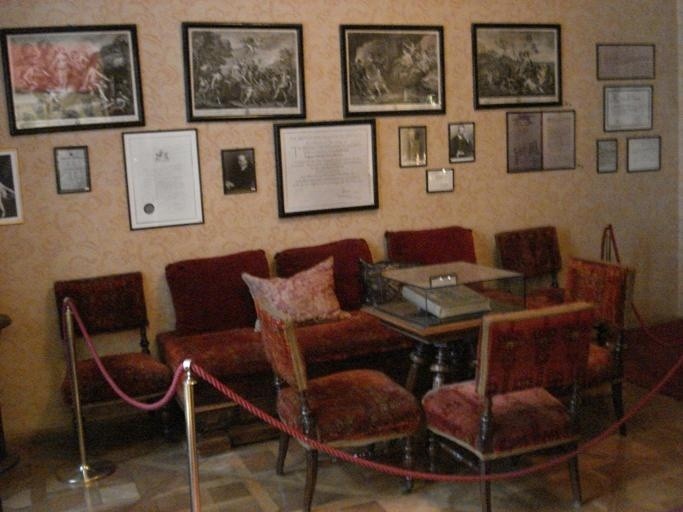}
    \end{minipage}
    \hfill
    \begin{minipage}[c]{0.78\textwidth}
        \textbf{Q:} What kind of room is this?? \\[0.3em]
        \textbf{\textit{\Nano}}: Museum room. \cmark\\
        \textbf{\textit{\Flash}}: This is a museum room. \cmark\\
        \textbf{\textit{nanoVLM}}: Living room. \xmark\\
        \textbf{\textit{LFM2.5-VL}}: This appears to be a formal meeting or conference room. The room features a long, plush red couch with matching chairs, creating a cozy yet professional atmosphere...\xmark
    \end{minipage}
    
    \vspace{0.6cm}
    \caption{\textbf{Qualitative comparison against sub-billion-parameter baselines.} Responses from \Nano and \Flash compared to nanoVLM-460M-8k and LFM2.5-VL-450M. The baselines frequently struggle with spatial reasoning, hallucinate scene details, or misidentify object counts and colors.}
    \label{fig:qualitative_vqa}
\end{figure*}

\cref{fig:qualitative_vqa} presents qualitative examples comparing \Nano\ and \Flash\ against sub-billion-parameter baselines (i.e., nanoVLM-460M-8k and LFM2.5-VL-450M) across diverse multimodal instructions. We observe that despite the substantial visual prefix compression in \Flash, it fully retains the reasoning capabilities of the full-token \Nano model. While their phrasing sometimes differs, both consistently provide accurate, grounded responses (indicated by \cmark). Crucially, we highlight that competitors frequently hallucinate details or fail simple spatial and counting tasks in these scenarios (marked with \xmark).

\subsection{Failure Cases and Limitations}
\label{app:qualitative_failures}
\begin{figure*}[htbp]
    \centering
    
    \begin{minipage}[c]{0.20\textwidth}
        \centering
        \includegraphics[width=\linewidth]{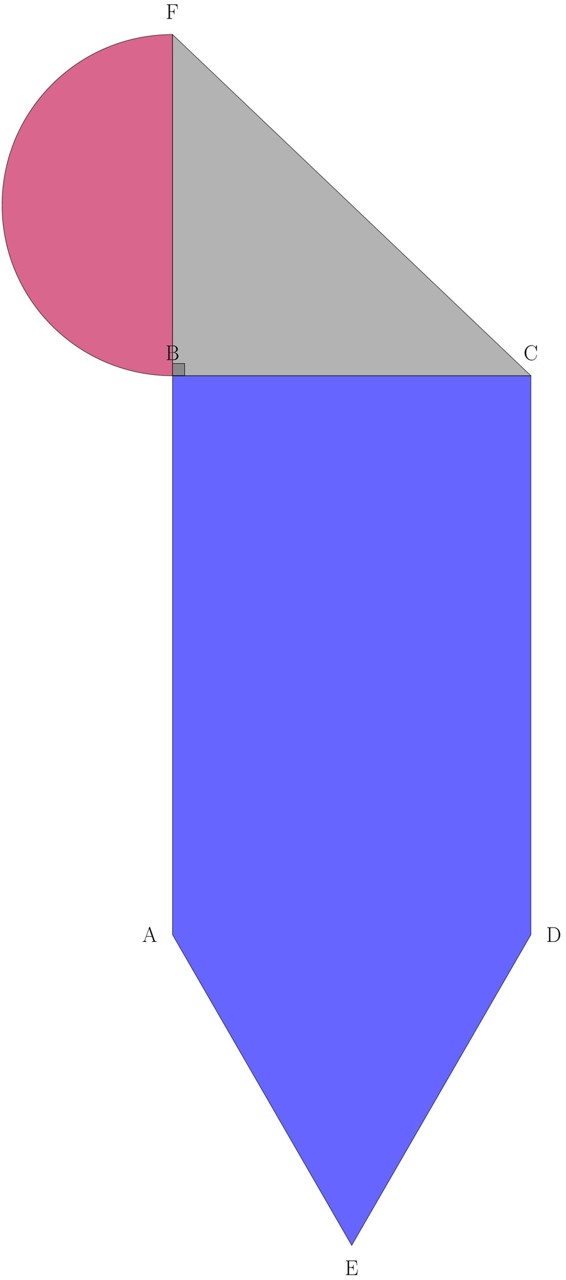}
    \end{minipage}
    \hfill
    \begin{minipage}[c]{0.78\textwidth}
        \textbf{Q:} The diagram shows the length BF = 14, the area of the gray triangle is 103, and the perimeter of the blue shape ABCDE is 90. Assuming BCDE forms a rectangle and ADE is an equilateral triangle, solve for the length of side AB. \hfill\GT{22.93}\\[0.3em]
        \textbf{\textit{\Nano}}: The answer is 22. \xmark\\
        \textbf{\textit{\Flash}}: The answer is 12. \xmark\\
        \textbf{\textit{nanoVLM}}: First, we find the area of the gray triangle using the formula for the area of a triangle given two sides and the included angle: (1/2) * base * height = 103. Since the base and height are equal, the area is 103. Then, we can find the area of the blue shape by subtracting the area of the gray triangle from the perimeter of the blue shape: 90 - 14 = 76. The area of the blue shape is 76, so the length of the AD side can be found by (76/2) * 2 = 14 * 2 = 28. Since ADE is an equilateral triangle, the length of the AE side is 28. The perimeter of the blue shape is 76, so the length of the AB side can be found by 76 - 28 - 28 = 28. Therefore the final answer is 28. \xmark\\
        \textbf{\textit{LFM2.5-VL}}: The length of the hypotenuse of the red triangle is 14 and the length of the side opposite to the degree of the angle marked with ``B'' is 14, so the degree of the angle marked with ``B'' equals $arcsin(\frac{14}{14}) = arcsin(1) = 90$. The area of the gray triangle is 103 and the length of the side opposite to the degree of the angle marked with ``B'' is 14, so the degree of the angle marked with ``B'' equals $arcsin(frac{103}{14}) = arcsin(7.43) = 45.53$. The length of the hypotenuse of the blue shape is 90 and the length of the side opposite to the degree of the angle marked with ``A'' is 7.43, so the degree of the angle marked with ``A'' equals $arcsin(frac{7.43}{90}) = arcsin(0.0828) = 5.88$. The length of the hypotenuse of the blue shape is 90 and the length of the side opposite to the degree of the angle marked with ``A'' is 5.88, so the degree of the angle marked with ``A'' equals $arcsin(frac{5.88}{90}) = arcsin(0.0652) = 3.89$. Therefore, the length of the side AB is equal to $90 * sin(3.89) = 90 * 0.0652 = 5.88$.\\Therefore, the answer is 5.88. \xmark
    \end{minipage}
    
    \vspace{0.4cm}
    \hrule
    \vspace{0.4cm}
    
    \begin{minipage}[c]{0.20\textwidth}
        \centering
        \includegraphics[width=\linewidth]{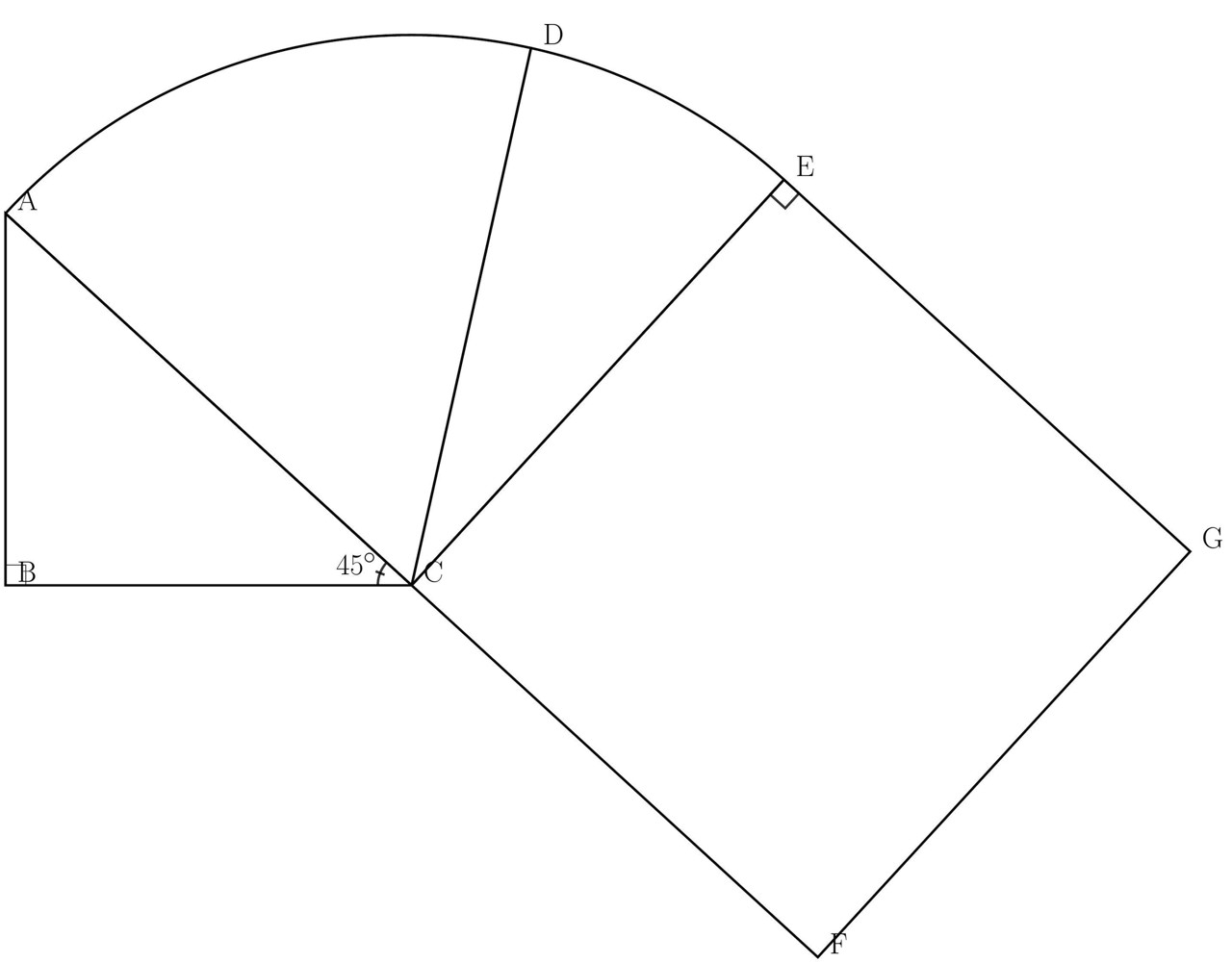}
    \end{minipage}
    \hfill
    \begin{minipage}[c]{0.78\textwidth}
        \textbf{Q:} List the letters labeling the vertices that have a right-angle symbol. \hfill\GT{B, E}\\[0.3em]
        \textbf{\textit{\Nano}}: B, C, G. \xmark\\
        \textbf{\textit{\Flash}}: C and G. \xmark\\
        \textbf{\textit{nanoVLM}}: The vertices that have a right angle symbol are the vertices that are corners of the rectangle. Therefore, the vertices that have a right angle symbol are D and E. \xmark\\
        \textbf{\textit{LFM2.5-VL}}: The vertices that have a right-angle symbol are A, C, and F. \xmark
    \end{minipage}
    
    \vspace{0.4cm}
    \hrule
    \vspace{0.4cm}
    
    \begin{minipage}[c]{0.20\textwidth}
        \centering
        \includegraphics[width=\linewidth]{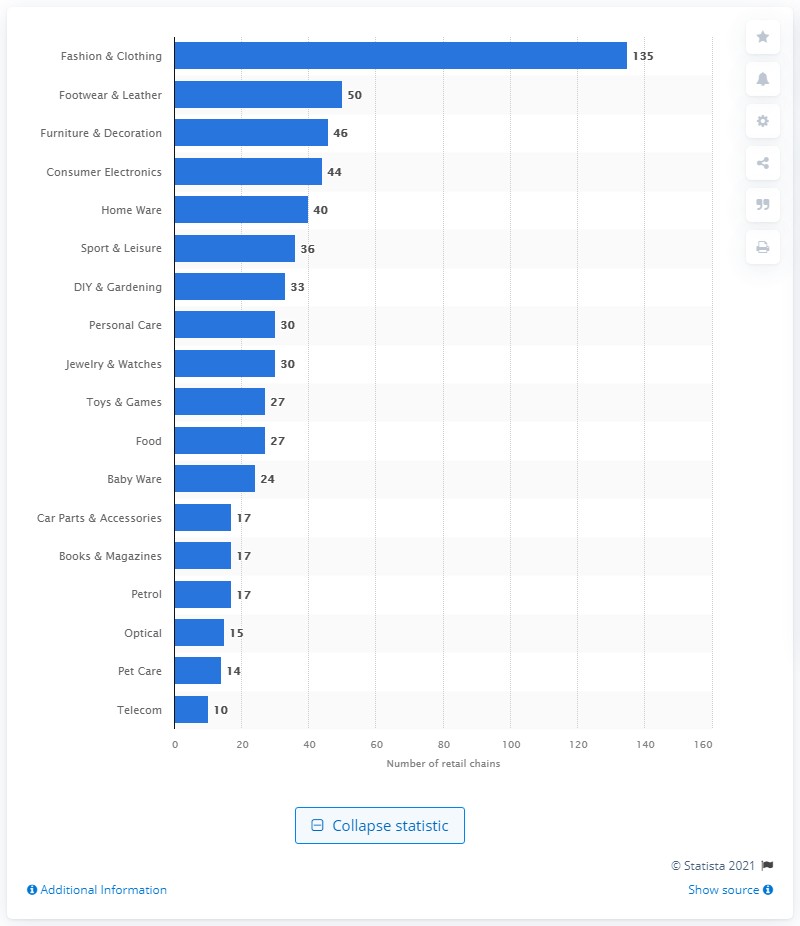}
    \end{minipage}
    \hfill
    \begin{minipage}[c]{0.78\textwidth}
        \textbf{Q:} Calculate the total number of retail chains for all categories that have fewer than 20 retail chains. Just give the number. \hfill\GT{90}\\[0.3em]
        \textbf{\textit{\Nano}}: 87. \xmark\\
        \textbf{\textit{\Flash}}: 46. \xmark\\
        \textbf{\textit{nanoVLM}}: 14. \xmark\\
        \textbf{\textit{LFM2.5-VL}}: 14. \xmark
    \end{minipage}
    
    \vspace{0.6cm}
    \caption{\textbf{Failure cases at the ${\sim}$0.5B scale.} Common failure modes across the entire sub-billion-parameter class. All evaluated models struggle with complex multi-step geometry (top), fine-grained symbol recognition (middle), and conditional data extraction from dense charts (bottom). Notably, while baseline models often generate verbose and hallucinated reasoning chains (e.g., LFM2.5-VL-450M calculating nonsensical $\arcsin$ values), our models tend to fail more concisely.}
    \label{fig:qualitative_failures}
\end{figure*}

To provide a transparent and comprehensive evaluation, \cref{fig:qualitative_failures} illustrates failure cases where all models in the sub-billion-parameter class struggle. At the ${\sim}0.5$B scale, certain tasks requiring intricate multi-step reasoning remain fundamentally challenging. Specifically, we observe systemic failures across all models on complex mathematical and geometric problem-solving, fine-grained symbol detection within diagrams, and conditional arithmetic over dense chart data. 

Interestingly, while the baselines (e.g., LFM2.5-VL-450M) often hallucinate verbose and mathematically nonsensical reasoning chains when failing, \Nano\ and \Flash\ typically fail concisely. These examples highlight the current boundaries of edge VLMs: while our diagnosis-driven recipe maximizes the available capacity and neutralizes doom loops, highly specialized multi-step capabilities generally benefit from the expanded capacity of larger models.

\subsection{Live On-Device Deployment}
\label{app:on_device_deployment}
\begin{figure*}[htbp]
    \centering
    \includegraphics[width=0.95\linewidth]{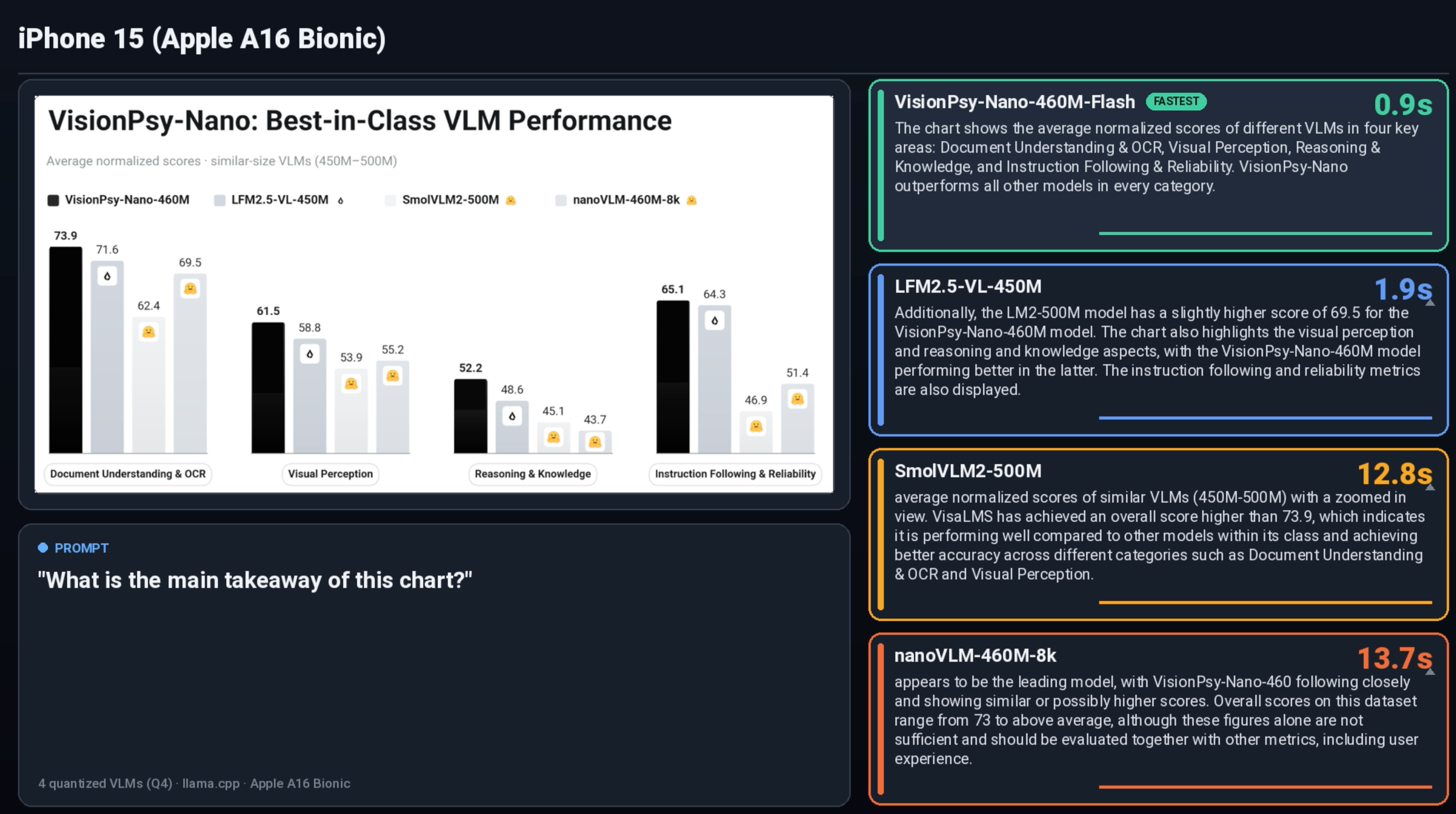}
    \caption{\textbf{Live on-device deployment (Apple A16 Bionic).} Inference on an iPhone 15 using \texttt{llama.cpp} (4-bit). \Flash achieves an interactive 0.9s time-to-first-token, completely bypassing the massive delays of default-policy models like nanoVLM-460M-8k (13.7s).}
    \label{fig:deployment_apple}
\end{figure*}

\begin{figure*}[htbp]
    \centering
    \includegraphics[width=0.95\linewidth]{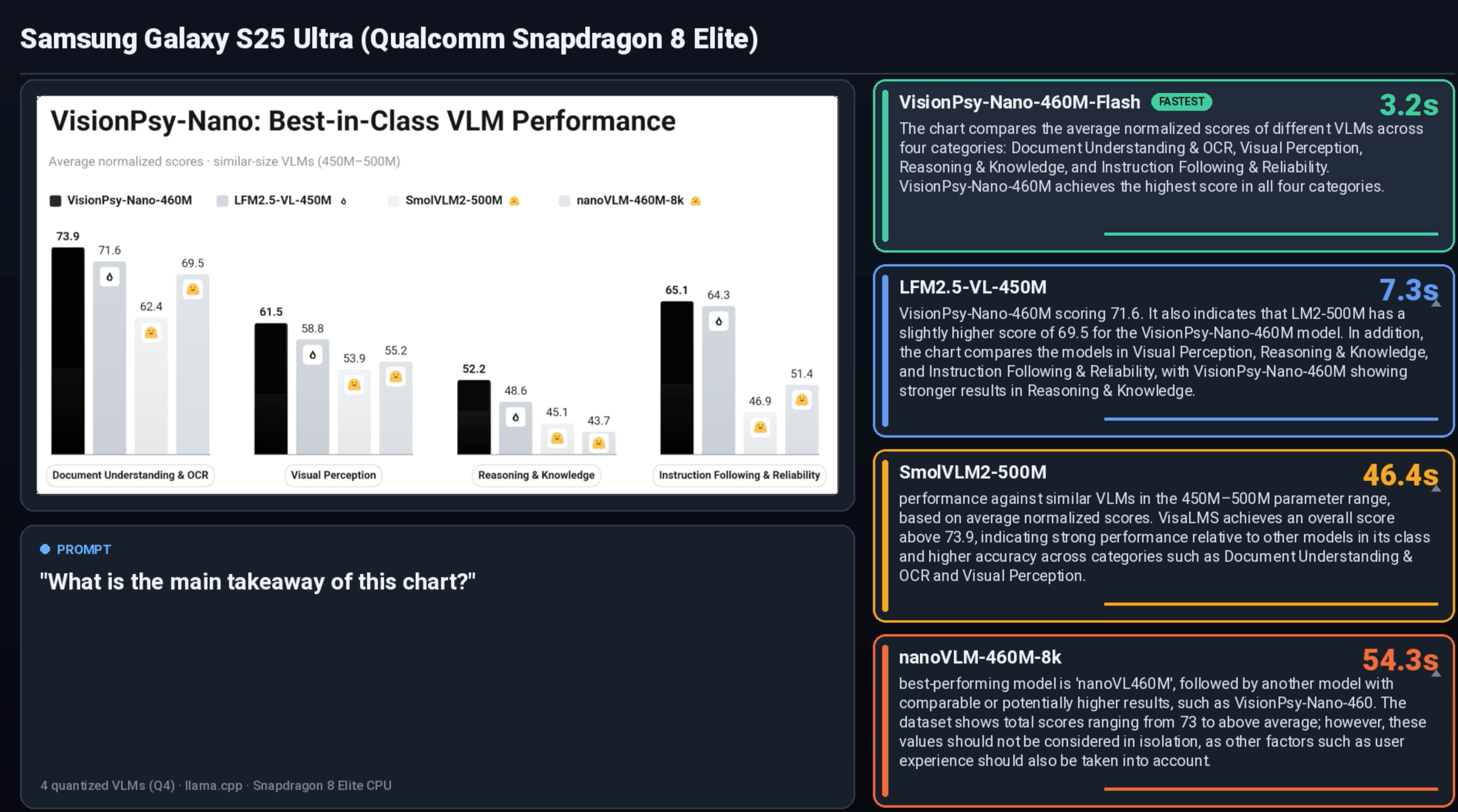}
    \caption{\textbf{Live on-device deployment (Qualcomm Snapdragon 8 Elite).} Inference on a Samsung Galaxy S25 Ultra. The visual prefix bottleneck affects even the newest mobile chips, with baselines taking up to 54.3s. \Flash reduces this delay to just 3.2s.}
    \label{fig:deployment_samsung}
\end{figure*}

\begin{figure*}[htbp]
    \centering
    \includegraphics[width=0.95\linewidth]{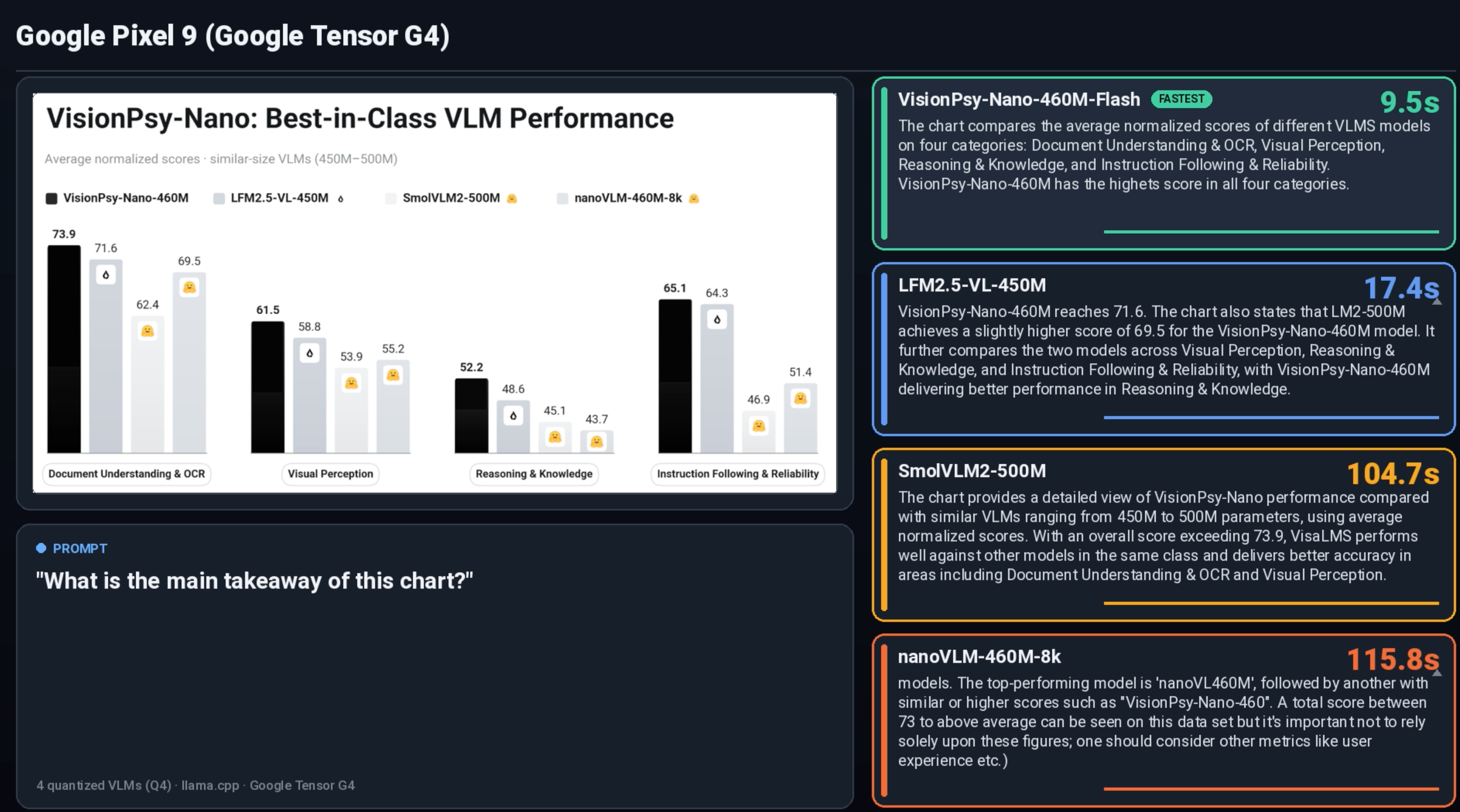}
    \caption{\textbf{Live on-device deployment (Google Tensor G4).} Inference on a Google Pixel 9. This represents the most constrained environment, where base models freeze for nearly two minutes (115.8s). \Flash successfully cuts this down to 9.5s, enabling usability.}
    \label{fig:deployment_google}
\end{figure*}

To visually demonstrate the practical impact of our token-bounding strategy across a diverse hardware ecosystem, we deployed the models on three flagship devices: an iPhone 15 (Apple A16 Bionic, \cref{fig:deployment_apple}), a Samsung Galaxy S25 Ultra (Qualcomm Snapdragon 8 Elite, \cref{fig:deployment_samsung}), and a Google Pixel 9 (Google Tensor G4, \cref{fig:deployment_google}). All models are executed under standard 4-bit weight quantization via \texttt{llama.cpp}.

Note that the time-to-first-token metrics displayed in these screenshots reflect a single, in-the-wild qualitative demonstration using the specific high-resolution chart shown in the UI. As such, they serve as a real-world showcase and are distinct from the rigorously controlled, standardized evaluation protocol (e.g., fixed resolutions, warm runs) documented in Sec.~\ref{app:on_device_protocol} and reported in our formal benchmarking tables.

The results reveal that the visual prefix bottleneck is a hardware-agnostic problem. Default-policy models like the base nanoVLM-460M-8k completely paralyze the interaction on all platforms, suffering from severe TTFT delays ranging from 13.7 seconds on the iPhone to a catastrophic 115.8 seconds on the Pixel 9. SmolVLM2-500M exhibits similar usability failures across the board. 

By contrast, \Flash consistently bypasses this prefill wall regardless of the underlying chip. It achieves sub-second responsiveness (0.9s) on the iPhone 15 and slashes the severe Pixel 9 delay down to just 9.5s, an over $12\times$ speedup compared to its full-token counterpart. This cross-platform validation visually confirms that our model successfully bridges the gap between theoretical benchmark scores and true, robust on-device interactivity.

\end{document}